\documentclass{article}
\usepackage{microtype}
\usepackage{graphicx}
\usepackage{subcaption}
\usepackage{booktabs} 

\usepackage{amsmath,amsfonts,bm}

\def\eqref#1{equation~\ref{#1}}

\def\1{\bm{1}}

\def\vs{{\bm{s}}}

\DeclareMathAlphabet{\mathsfit}{\encodingdefault}{\sfdefault}{m}{sl}
\SetMathAlphabet{\mathsfit}{bold}{\encodingdefault}{\sfdefault}{bx}{n}

\usepackage[most]{tcolorbox}
\usepackage{hyperref}
\usepackage[table]{xcolor}
\definecolor{original}{gray}{0.80}
\definecolor{our}{gray}{0.65}
\definecolor{best}{RGB}{252, 230, 213}     
\definecolor{second}{RGB}{230, 230, 250} 
\definecolor{emphasize}{RGB}{226, 140, 217} 

\usepackage[preprint]{icml2026}
\usepackage{array}
\usepackage{amsmath}
\usepackage{amssymb}
\usepackage{mathtools}
\usepackage{amsthm}
\usepackage{pifont}
\usepackage{bm}
\usepackage[capitalize,noabbrev]{cleveref}

\theoremstyle{plain}

\theoremstyle{definition}

\theoremstyle{remark}

\usepackage{multirow}
\usepackage{makecell}
\usepackage{enumerate}
\usepackage{enumitem}
\usepackage{pifont}
\usepackage{xcolor}
\usepackage{xurl}
\renewcommand{\algorithmiccomment}[1]{/ / #1}
\usepackage[textsize=tiny]{todonotes}

\def\eg{\emph{e.g.}}
\def\ie{\emph{i.e.}}

\def\vs{\emph{vs.}}

\def\fig{Fig. }
\def\tab{Tab. }
\def\sec{Sec. }
\def\alg{Alg. }

\newcommand{\benchmark}[1]{OmniVCHall}
\newcommand{\approach}[1]{TriCD}

\icmltitlerunning{Learning to Decode Against Compositional Hallucination in Video Multimodal Large Language Models}

\begin{document}

\twocolumn[
  \icmltitle{Learning to Decode Against Compositional Hallucination in \\Video Multimodal Large Language Models}
  \icmlsetsymbol{equal}{*}

  \begin{icmlauthorlist}
    \icmlauthor{Wenbin Xing}{sysu}
    \icmlauthor{Quanxing Zha}{hqu}
    \icmlauthor{Lizheng Zu}{szu}
    \icmlauthor{Mengran Li}{sysu}
    \icmlauthor{Ming Li}{gm}
    \icmlauthor{Junchi Yan}{sjtu}
  \end{icmlauthorlist}

  \icmlaffiliation{sysu}{Sun Yat-sen University}
  \icmlaffiliation{hqu}{Huaqiao University}
  \icmlaffiliation{szu}{Shenzhen University}
  \icmlaffiliation{gm}{Guangming Laboratory}
  \icmlaffiliation{sjtu}{Shanghai Jiao Tong University}

  \icmlcorrespondingauthor{Ming Li}{ming.li@u.nus.edu}

  \icmlkeywords{Cross-modal video understanding, Contrastive decoding, Model capability}

  \vskip 0.3in
]



\printAffiliationsAndNotice{}  

\begin{abstract}
\vspace{-0.5em}
Current research on video hallucination mitigation primarily focuses on isolated error types, leaving \textit{compositional} hallucinations—arising from incorrect reasoning over multiple interacting spatial and temporal factors largely underexplored. We introduce \textbf{\benchmark{}}, a benchmark designed to systematically evaluate both isolated and compositional hallucinations in video multimodal large language models (VLLMs). \benchmark{} spans diverse video domains, introduces a novel camera-based hallucination type, and defines a fine-grained taxonomy, together with adversarial answer options (\eg, “All are correct” and “None of the above”) to prevent shortcut reasoning. The evaluations of 39 representative VLLMs reveal that even advanced models (\eg, Qwen3-VL and GPT-5) exhibit substantial performance degradation.
We propose \textbf{\approach{}}, a contrastive decoding framework with a triple-pathway calibration mechanism. An adaptive perturbation controller dynamically selects distracting operations to construct negative video variants, while a saliency-guided enhancement module adaptively reinforces grounded token-wise visual evidences. These components are optimized via reinforcement learning to encourage precise decision-making under compositional hallucination settings.
Experimental results show that \approach{} consistently improves performance across two representative backbones, achieving an average accuracy improvement of over 10\%. The data and code can be find at \url{https://github.com/BMRETURN/OmniVCHall}.
\vspace{-0.5em}
\end{abstract}

\section{Introduction}
\vspace{-0.5em}
The rapid evolution of Video Large Language Models (VLLMs) has enabled unprecedented capabilities in video understanding \cite{zou2024seconds, tang2025video, Qwen3-VL, google2025gemini3pro}. Despite this progress, the reliability of these models remains compromised by pervasive hallucinations, manifested as generated content that is inconsistent with the underlying visual evidence \cite{rawte2023survey, sahoo2024comprehensive}. A growing body of work has investigated this issue and proposed methods to mitigate hallucinations in VLLMs \cite{bansal2024videocon, wang2024videohallucer, zhang2024eventhallusion, kong2025mhbench, rawal2025argus}. They predominantly focus on single or isolated hallucination types, targeting specific failure modes such as action misinterpretation \cite{kong2025mhbench} or temporal inconsistencies \cite{choong2024vidhal, li2025vidhalluc}.

However, in real-world settings, diverse hallucination types often emerge simultaneously and interact in a highly coupled manner  \cite{li2023evaluating, maaz2024video, wang2024vigc, wang2024videohallucer, bae2025mash, yang2025discovering, chytas2025reco}. This failure mode is referred as \textit{compositional} hallucinations, which is particularly pronounced in VLLMs, as input videos typically contain rich and interdependent spatial and temporal cues. For example, object attributes, motion patterns, and camera dynamics are often jointly involved within a single query. Despite their prevalence, compositional hallucinations remain largely underexplored in existing cross-modal video studies. 

\begin{figure}[t!]
\centering
\includegraphics[width=0.95\linewidth]{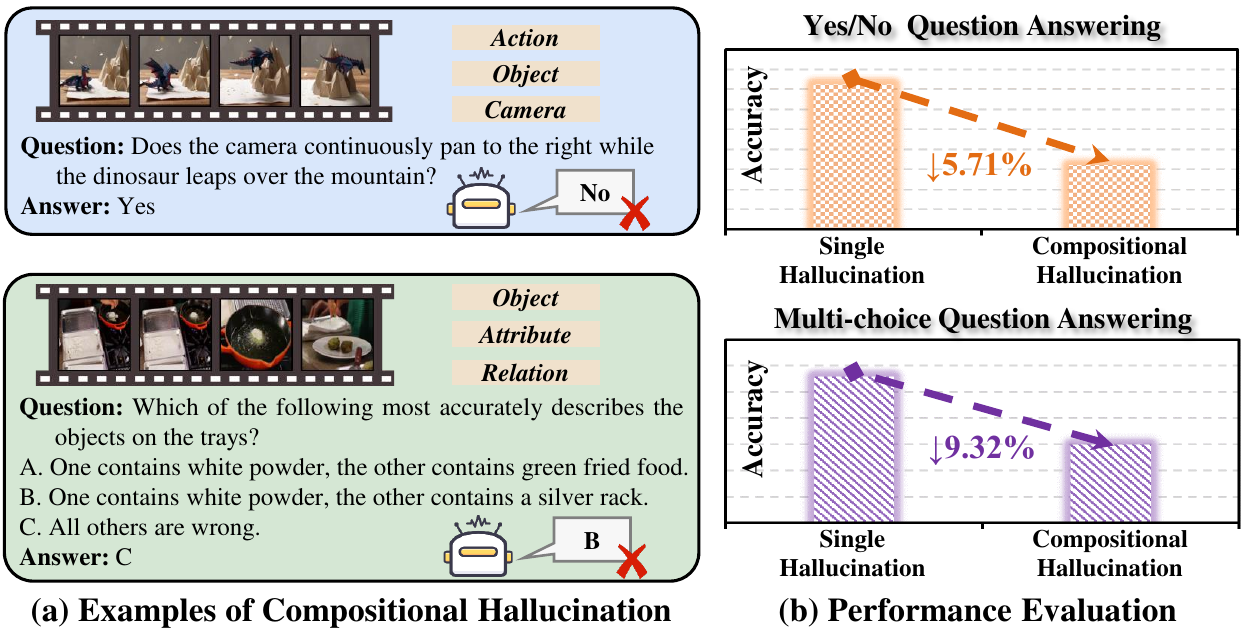}
\vspace{-0.5em}
\caption{(a) Existing VLLMs frequently struggle with cross-modal video understanding tasks involving compositional hallucinations.
(b) Extensive evaluation across 39 models reveals a pronounced accuracy drop ($\downarrow$5.71\% and $\downarrow$9.32\%) when transitioning from single-factor hallucination queries to compositional ones.}
\label{fig_motivation}
\vspace{-1.5em}
\end{figure}

To systematically address this gap, we introduce \benchmark{}, a comprehensive benchmark specifically designed to evaluate both isolated and compositional hallucinations. Unlike prior datasets that rely on coarse taxonomies, \benchmark{} defines eight fine-grained hallucination types, including a novel camera type that exposes observer-perspective vulnerabilities largely overlooked in existing studies. In addition, \benchmark{} incorporates adversarial options (\eg, “All others are correct” and “None of the above”) to discourage shortcut reasoning and enforce explicit grounding in the visual evidence.
In total, \benchmark{} comprises 823 high-quality videos, drawn from both authentic real-world footage and carefully curated AI-generated content, paired with 9,027 visual question answering (VQA) samples. \fig \ref{fig_motivation} (a) provides two representative examples of VLLMs failing on yes/no and multiple-choice VQA tasks involving compositional hallucination. \fig \ref{fig_motivation} (b) further illustrates that models achieve significantly lower accuracy on queries involving compositional hallucinations compared to those containing a single type. These results indicate that while VLLMs may handle isolated inconsistencies, their reasoning robustness collapses when confronted with multi-factor compositional contradictions.

To tackle this issue, we introduce \approach{}, a triple-pathway video contrastive decoding calibration framework that unifies negative suppression via context-aware perturbation selection with saliency-aware spatiotemporal enhancement.
Specifically, we propose an adaptive perturbation controller that dynamically selects context-aware video suppression operations, enabling flexible and instance-specific negative distraction.
In parallel, we develop a saliency-guided enhancement module that strengthens visual evidence by fusing DINOv3 \cite{simeoni2025dinov3} spatial features with Farneback \cite{farneback2003two} temporal motion cues, thereby improving visual grounding for complex spatiotemporal reasoning.
Both modules are optimized via reinforcement learning to align the calibration process with the unique reasoning demands of each query.
Extensive experimental results show that \approach{} achieves substantial accuracy gains, \ie, 9.61\% for Qwen3-VL-Instruct-8B and 12.03\% for VideoLLaMA3-7B, demonstrating its superiority as a robust and scalable solution for mitigating compositional hallucinations in VLLMs. Our contributions are:
\begin{itemize}[leftmargin=*]
\vspace{-1.0em}
    \item \textbf{Compositional Hallucinations in VLLMs.}
    We empirically show that queries involving multiple interacting hallucination factors pose substantially greater challenges than isolated hallucination types. To the best of our knowledge, this is the first systematic study of compositional hallucinations in VLLMs.

    \item \textbf{Benchmark.}
    We introduce \benchmark{}, a comprehensive benchmark comprising high-quality videos, VQA pairs, a novel hallucination category, and adversarial answer options designed to prevent shortcut reasoning, enabling systematic evaluation of compositional hallucinations in existing VLLMs.

    \item \textbf{Method.}
    We propose \approach{}, a triple-pathway contrastive decoding framework that combines adaptive negative suppression with saliency-aware spatiotemporal enhancement. Without updating the VLLM parameters, \approach{} consistently improves performance in mitigating compositional hallucinations.
    \vspace{-1.0em}
\end{itemize}

\begin{figure*}[t]
\centering
\includegraphics[width=0.90\linewidth]{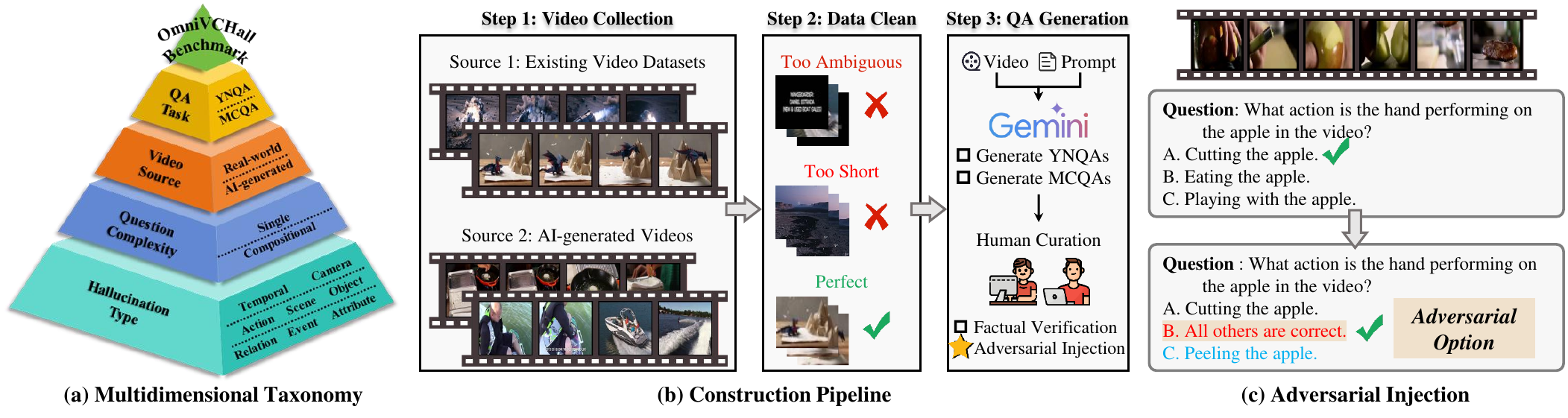}
\vspace{-0.5em}
\caption{Overview of the \benchmark{} benchmark. (a) shows a hierarchical structure. (b) shows a three-step pipeline. (c) utilizes adversarial answer options (\eg, ``All are correct” and ``None of the above”) to discourage shortcut reasoning.}
\label{fig_benchmark}
\vspace{-1.0em}
\end{figure*}

\section{\benchmark{} Benchmark}
Existing video hallucination benchmarks are often confined to narrow categories and predominantly focus on single-dimensional queries \cite{choong2024vidhal, kong2025mhbench, li2025vidhalluc}. However, in real-world scenarios, hallucinations rarely occur in isolation. They tend to emerge as entangled errors spanning multiple facets of video content. To bridge this gap, this section introduces \benchmark{}, a comprehensive benchmark designed to systematically evaluate both single and compositional hallucinations in VLLMs.

\subsection{Taxonomy Design}\label{Taxonomy}
\vspace{-0.5em}
As illustrated in \fig \ref{fig_benchmark} (a), \benchmark{} establishes a structured taxonomy that evaluates VLLMs across four critical perspectives,  ensuring a rigorous assessment.

\vspace{-0.8em}
\paragraph{Fine-grained Hallucination Type.}
At the foundational level, we define eight fine-grained hallucination types covering distinct semantic facets of video understanding. Among them, we introduce a camera-related hallucination type to explicitly evaluate hallucination induced by cinematic dynamics, where camera operations such as zooming or panning are mistakenly interpreted as physical object motion. We formally define each type in \tab \ref{tab_definition_} and provide representative examples in \sec \ref{Hallucination Type}. This fine-grained taxonomy enables a targeted analysis of model vulnerabilities.

\begin{table}[tb!]
  \centering
  \caption{Definition of eight hallucination types.}
  \label{tab_definition_}
  \vspace{-0.5em}
  \resizebox{0.48\textwidth}{!}{
  \begin{tabular}{cl} 
    \toprule
    \textbf{Type} & \textbf{Definition} \\
    \midrule
    Object & Presence or identity of physical entities.  \\
    Scene  & The overarching environment or setting. \\
    Event & High-level semantic units or causal occurrences. \\
    Action & Physical movement or behavioral patterns. \\
    Relation  & Spatial or logical interactions between entities. \\
    Attribute  & Static properties like color, size, or material. \\
    Temporal  & Chronological order or duration of occurrences. \\
    Camera & Perception of cinematic lens dynamics. \\
    \bottomrule
  \end{tabular}
  }
  \vspace{-0.5em}
\end{table}

\vspace{-0.8em}
\paragraph{Hierarchical Question Complexity.}
\benchmark{} incorporates two levels of task complexity. Single hallucination (S) tasks focus on an isolated error type, whereas compositional hallucination (C) tasks involve queries that span multiple hallucination types simultaneously, thereby exposing fundamental reasoning bottlenecks in current VLLMs.

\vspace{-0.8em}
\paragraph{Diverse Video Source.}
\benchmark{} comprises both real-world and AI-generated videos to ensure source diversity, capturing hallucination behaviors under both natural visual dynamics and synthetic content distributions.

\vspace{-0.8em}
\paragraph{Multi-faceted QA Task.}
To provide a comprehensive evaluation of model responses, \benchmark{} adopts two primary task formats: binary yes/no question answering (YNQA) for probing factual verification, and multi-choice question answering (MCQA) for assessing discriminative reasoning among competing semantic candidates. By intersecting these task formats with varying question complexity, we organize the evaluation samples into four sub-tasks:

\begin{itemize}[nosep, leftmargin=1.5em]
\item S\_YNQA: Binary factual verification on a single type.
\item C\_YNQA: Binary factual verification on multiple types.
\item S\_MCQA: Discriminative reasoning on a single type.
\item C\_MCQA: Discriminative reasoning on multiple types.
\end{itemize}

\subsection{Benchmark Construction}\label{Construction}
\vspace{-0.5em}
The construction of \benchmark{} follows a carefully designed three-step pipeline to ensure both data diversity and annotation reliability, as illustrated in \fig \ref{fig_benchmark} (b).

\vspace{-0.8em}

\paragraph{Video Collection and Curation.}
We initially curated a pool of 1,000 candidate videos, which were manually reviewed and filtered to ensure that each video provides clear and unambiguous visual evidence for reliable hallucination detection. This process yields a carefully curated corpus of 823 high-quality video clips from 11 heterogeneous domains, encompassing authentic real-world scenarios as well as synthetically generated videos with complex artifacts. Detailed source distributions are provided in \sec \ref{Data Sources}.

\vspace{-0.8em}
\paragraph{QA Generation with Adversarial Injection.} 
We adopt a hybrid human-AI approach for QA generation. To establish a reliable semantic anchor, we first employ manual annotation to produce detailed ground-truth captions for each video clip, accurately documenting the underlying spatiotemporal dynamics and object interactions. These human-verified descriptions serve as stable textual references, effectively reducing the risk of model-induced noise during subsequent QA construction. Secondly, Gemini-2.5-Pro \cite{google2025gemini2_5pro} is utilized to generate initial candidate YNQA and MCQA questions based on the videos, human-annotated captions, and structured prompts. All generated questions are subsequently manually curated by expert annotators to verify factual correctness and to ensure that each compositional query is annotated with the corresponding hallucination types for all relevant semantic facets. A notable characteristic of \benchmark{} is the dynamic injection of adversarial options in MCQA tasks. Expert annotators manually incorporate options (\eg, “All are correct” and “None of the above”) to discourage reliance on simple elimination strategies or language-only priors. This encourages performing holistic cross-modal verification of each candidate before selecting an answer. Relevant cases are provided in \sec \ref{Adversarial Option}. To quantify the quality of the curated dataset and establish an empirical upper bound, we conduct a human performance evaluation involving three independent evaluators. As shown in \tab \ref{tab_benchmark_results}, human evaluators achieve an average accuracy of 0.95, significantly outperforming current state-of-the-art VLLMs. Human performance remains consistently high and relatively stable across all sub-categories and task complexities. This indicates that, although adversarial options increase the difficulty for models, the tasks remain well-defined and solvable for human reasoners.

\begin{figure}[!b]
\centering
\includegraphics[width=0.85\linewidth]{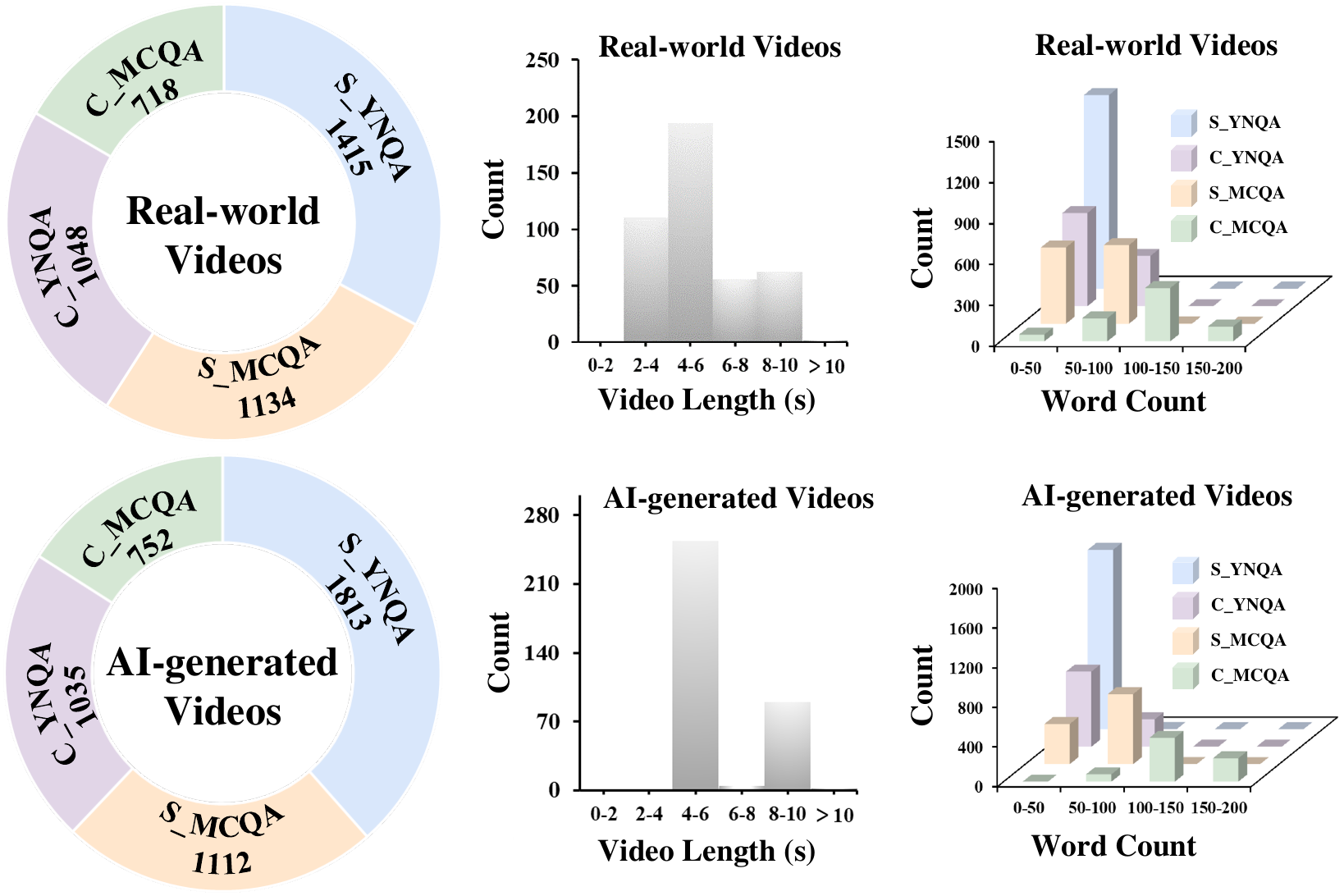}
\vspace{-0.5em}
\caption{Statistical analysis of \benchmark{}. From left to right: distribution of the four sub-tasks across domains, histograms of video durations, and word count distributions highlighting the complexity of compositional tasks.}
\label{fig_statistic}
\end{figure}

\begin{table*}[t!]
  \centering
  \caption{Comparison of \benchmark{} with other video hallucination datasets. \colorbox{our}{Ours} provides the most comprehensive coverage across eight hallucination types, including the newly defined camera type, while uniquely balancing real-world and AI-generated sources.}
  \vspace{-0.5em}
  \label{tab_other_datasets}
  \setlength{\tabcolsep}{1.5pt}
  \resizebox{0.98\textwidth}{!}{ 
  \begin{tabular}{lcccccccccccccc} 
    \toprule
    \multirow{2}{*}{\textbf{Benchmark}} & \multicolumn{8}{c}{\textbf{Hallucination Type}} & \multicolumn{3}{c}{\textbf{QA}} & \multicolumn{3}{c}{\textbf{Video}} \\
    \cmidrule(lr){2-9} \cmidrule(lr){10-12} \cmidrule(lr){13-15}
    & Object & Scene & Event & Action & Relation & Attribute & Temporal & Camera & YN & MC & Count & Real & Generated & Count \\
    \midrule
TempCompass \cite{liu2024tempcompass} & \ding{55} & \ding{55} & \ding{51} & \ding{51} & \ding{51} & \ding{51} & \ding{51} & \ding{55} & \ding{51} & \ding{51} & 7,540 & \ding{51} & \ding{55} & 410 \\
VidHal \cite{choong2024vidhal} & \ding{51} & \ding{55} & \ding{55} & \ding{51} & \ding{51} & \ding{51} & \ding{51} & \ding{55} & \ding{55} & \ding{51} & 1,000 & \ding{51} & \ding{55} & 1,000 \\
VideoHallucer \cite{wang2024videohallucer} & \ding{51} & \ding{55} & \ding{51} & \ding{55} & \ding{51} & \ding{55} & \ding{51} & \ding{55} & \ding{51} & \ding{55} & 1,800 & \ding{51} & \ding{55} & 948 \\
EventHallusion \cite{zhang2024eventhallusion} & \ding{55} & \ding{55} & \ding{51} & \ding{55} & \ding{55} & \ding{55} & \ding{55} & \ding{55} & \ding{55} & \ding{55} & 711 & \ding{51} & \ding{55} & 400 \\
MHBench \cite{kong2025mhbench} & \ding{55} & \ding{55} & \ding{55} & \ding{51} & \ding{55} & \ding{55} & \ding{55} & \ding{55} & \ding{51} & \ding{51} & 3,600 & \ding{51} & \ding{55} & 1,200 \\
VidHalluc \cite{li2025vidhalluc} & \ding{55} & \ding{51} & \ding{55} & \ding{51} & \ding{55} & \ding{55} & \ding{51} & \ding{55} & \ding{51} & \ding{51} & 9,295 & \ding{51} & \ding{55} & 5,002 \\
ELV-Halluc \cite{lu2025elv} & \ding{51} & \ding{55} & \ding{51} & \ding{51} & \ding{55} & \ding{55} & \ding{51} & \ding{55} & \ding{51} & \ding{55} & 4,800 & \ding{51} & \ding{55} & 200 \\
    \midrule
\rowcolor{original} \textbf{\benchmark{} (Ours)} & \ding{51} & \ding{51} & \ding{51} & \ding{51} & \ding{51} & \ding{51} & \ding{51} & \ding{51} & \ding{51} & \ding{51} & 9,027 & \ding{51} & \ding{51} & 823 \\
    \bottomrule
  \end{tabular}
  }
  \vspace{-0.5em}
\end{table*}

\subsection{Data Statistics}\label{Statistics}
\vspace{-0.5em}
As illustrated in \fig \ref{fig_statistic}, we conduct a quantitative analysis of \benchmark{} across multiple dimensions to demonstrate its diversity and complexity. The benchmark contains 823 videos, balanced between real-world (423) and AI-generated (400) domains, resulting in a total of 9,027 queries. Temporally, most videos span 2 to 10 seconds, ensuring sufficient dynamic content for hallucination assessment. A key distinction of \benchmark{} lies in the higher linguistic complexity of compositional tasks. Compositional-type queries contain significantly more words than single-type queries, indicating the sophisticated reasoning required to disentangle multiple hallucination types. \tab \ref{tab:omnivh_adversarial} presents the distribution of adversarial options in \benchmark{}, which is designed to mitigate shortcut reasoning by requiring models to perform explicit cross-modal verification for each candidate. The query distributions across different hallucination types are detailed in \sec \ref{QA Distribution}.

Inter-Annotator Agreement (IAA) \cite{landis1977measurement} analysis confirms the high reliability of \benchmark{}, with Cohen’s Kappa scores of 0.84 for video description and 0.81 for QA validity (see in \sec \ref{sec:iaa}).

\begin{table}[tb!]
  \centering
  \caption{Statistics of adversarial option injection in \benchmark{}. “AD” denotes queries with injected adversarial options.}
  \vspace{-0.5em}
  \label{tab:omnivh_adversarial}
  \resizebox{0.43\textwidth}{!}{ 
  \begin{tabular}{ccccc} 
    \toprule
    \multirow{2}{*}{\textbf{Type}} & \multicolumn{2}{c}{\textbf{Real-world}} & \multicolumn{2}{c}{\textbf{AI-generated}} \\
    \cmidrule(lr){2-5}
    & S\_MCQA & C\_MCQA & S\_MCQA & C\_MCQA \\
    \midrule
    AD  & 315  & 243 & 268  & 179 \\ 
    Total  & 1134 & 718 & 1112 & 752 \\
    Ratio & 27.78\% & 33.84\% & 24.10\% & 23.80\% \\
    \bottomrule
  \end{tabular}
  }
  \vspace{-0.5em}
\end{table}

Finally, as summarized in \tab \ref{tab_other_datasets}, \benchmark{} offers a more comprehensive evaluation of video-text misalignment by introducing the novel camera type hallucination and maintaining a balanced mix of real-world and AI-generated videos. This benchmark provides a valuable testbed for the community to develop more reliable and robust VLLMs.

\section{\approach{} Framework}
\begin{figure*}[htbp]
\centering
\includegraphics[width=0.85\linewidth]{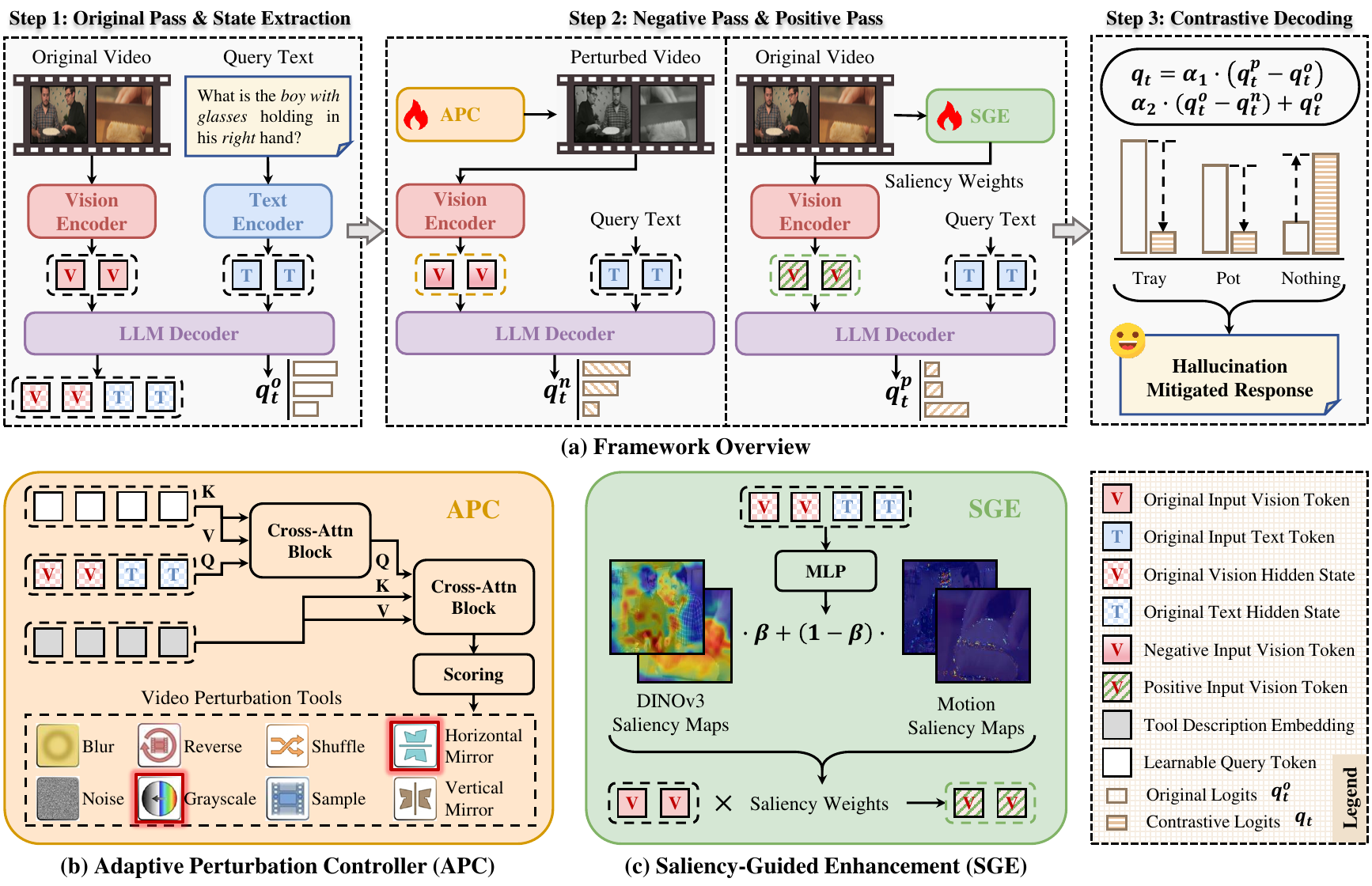}
\vspace{-0.5em}
\caption{\approach{} framework. (a) employs a three-step process that uses contrastive decoding to refine final predictions. (b) dynamically selects the most contextually relevant tools from a bank of eight video perturbation tools via cross-attention to construct a negative sample. (c) fuses spatial (DINOv3) and temporal (motion) saliency maps to reweight vision tokens, anchoring the positive pass to critical evidence.}
\label{fig_framework}
\vspace{-1.0em}
\end{figure*}

\vspace{-0.5em}
Existing Contrastive Decoding (CD) methods rely on static perturbations (\eg, shuffling) that fail to address the compositional hallucinations prevalent in complex video reasoning \cite{zhang2024eventhallusion, kong2025mhbench}. Effective mitigation requires semantically adaptive negative samples, dynamically tailored to the video-query context. However, suppression alone is insufficient as it lacks explicit guidance toward critical visual evidence \cite{li2025vidhalluc}. To resolve this, this section introduces \approach{}, a three-step framework coupling an Adaptive Perturbation Controller (APC) with Saliency-Guided Enhancement (SGE), as shown in \fig \ref{fig_framework}~(a).

\subsection{Original Pass \& State Extraction}\label{Original}
\vspace{-0.5em}
\approach{} begins by establishing a baseline understanding of the video-query context to guide subsequent adaptive perturbations and saliency enhancements. For an input video $V$ and textual query $T$, visual and textual tokens ($\mathbf{X}_v, \mathbf{X}_t$) are concatenated for the LLM decoder. We extract hidden states $\mathbf{H}$ from the final transformer layer as a semantic anchor representing the grounded video-query context. For a VLLM (defined as $\theta$), the predicted original logit $q_t^o$ at $t$ is:
\begin{equation}
q_t^o = \text{logit}{\theta} (y_t \mid V, T, y_{<t}),
\end{equation}
where $y_{<t}$ are previous tokens. These original logits represent the initial, uncalibrated probability distribution over vocabulary $\mathcal{V}$ and remain susceptible to hallucinations.

\subsection{Negative Pass \& Positive Pass}\label{Negative_Positive}
\vspace{-0.5em}
\approach{} executes two parallel branches to establish a calibrated inference boundary. The APC module generates context-aware negative samples to define what the model should avoid. The SGE branch anchors the model to evidence-rich regions to define what it should prioritize.

\subsubsection{Adaptive Perturbation Controller}
\vspace{-0.5em}
To address compositional hallucinations that static methods fail to capture, the APC module dynamically selects the most effective perturbation $\tau$ from a dictionary of eight tools $\mathcal{P}$. Each tool is designed to disrupt specific visual or temporal cues, providing a diverse set of negative samples for the contrastive decoding process. Their descriptions and examples are detailed in \sec \ref{Tool Analysis} and \sec \ref{Visual Examples}.

The APC module employs a dual-stage cross-attention mechanism to identify contextually relevant negative tools, as illustrated in \fig \ref{fig_framework} (b). Initially, learnable query tokens $\mathbf{Q}$ interact with the video-query hidden states $\mathbf{H}$ to encode specific semantic context. These conditioned queries then attend to the tool description embeddings $\mathbf{D}_{tool}$, which are derived by extracting the VLLM's hidden states after processing text tokens that define the functionality of each perturbation tool. Specifically, for each tool, we construct the comprehensive analysis, as detailed in \sec \ref{Tool Analysis}. 

Following the scoring head, tools are ranked in descending order, and all tools within a cumulative probability threshold $\gamma$ are selected to produce the perturbed sample $V^{-}$. Crucially, $V^{-}$ is designed to intentionally amplify the model's reliance on linguistic priors and hallucinations by corrupting critical spatiotemporal grounding. The resulting negative logit $q_t^n$ captures these induced hallucination pathways:
\begin{equation}
q_t^n = \text{logit}_{\theta} (y_t \mid V^{-}, T, y_{<t}),
\end{equation}
thereby serving as an informative “lower bound” that enables the subsequent contrastive decoding step to filter out misleading signals. Details are provided in \sec \ref{appendix_apc}.

\subsubsection{Saliency-Guided Enhancement}
\vspace{-0.5em}
Complementary to the negative pass in APC, the SGE module aims to sharpen the positive visual signal by anchoring the model's attention to the most informative spatiotemporal regions. SGE fuses object-centric spatial saliency with motion-aware temporal saliency via a learnable gating mechanism. Their examples are shown in \sec \ref{Visual Examples}.

\vspace{-0.8em}
\paragraph{Dual Saliency Extraction.}
Following \cite{li2025vidhalluc}, we extract spatial saliency $\mathbf{s}_{spa}$ by computing the attention weights from the [CLS] token to all patch tokens in the last layer of DINOv3 \cite{simeoni2025dinov3}, effectively capturing foreground object importance. Simultaneously, we estimate temporal saliency $\mathbf{s}_{mot}$ using the Farneback algorithm \cite{farneback2003two} to calculate optical flow. To isolate intentional actions from background jitter, we apply a dual-stage filtering process (Gaussian smoothing and temporal band-pass filtering) to the raw motion magnitude.

\vspace{-0.8em}
\paragraph{Dynamic Fusion.}
As illustrated in \fig \ref{fig_framework} (c), the final spatiotemporal signal is modulated by a learnable gating network. Conditioned on the context-rich hidden states $\mathbf{H}$, an MLP predicts a dynamic fusion weight $\beta \in [0, 1]$, yielding the unified weight: $\mathbf{w}_{sal} = \beta \cdot \mathbf{s}_{spa} + (1-\beta) \cdot \mathbf{s}_{mot}$. The positive vision tokens $\mathbf{X}_v'$ are then enhanced via element-wise reweighting, $\mathbf{X}_v' = \mathbf{w}_{sal} \odot \mathbf{X}_v$, and fed into the decoder to obtain the positive logit: 
\begin{equation}
q_t^p = \text{logit}_{\theta} (y_t \mid \mathbf{X}_v', \mathbf{X}_t, y_{<t}).
\end{equation}
This mechanism establishes a grounded “upper bound” by ensuring the model remains anchored to both critical objects and their dynamic interactions. Technical implementation details are provided in \Cref{appendix_sge}.

\subsection{Triple-Pathway Contrastive Decoding}\label{Contrastive Decoding}
\vspace{-0.5em}
With the original logits $q_t^o$, the adaptive negative logits $q_t^n$, and the saliency-guided positive logits $q_t^p$ successfully extracted, \approach{} integrates them into a unified calibration framework. By treating the original prediction as a baseline and applying directed residual corrections, the model effectively widens the margin between grounded visual evidence and potential hallucination pathways. The final calibrated logit $q_t$ is formulated as below:
\begin{equation}
q_t = q_t^o + {\alpha}_1 \cdot (q_t^p - q_t^o) + {\alpha}_2 \cdot (q_t^o - q_t^n),
\label{ours}
\end{equation}
where both $\alpha_1$ and $\alpha_2$ are hyperparameters that adjusts the penalty strengt. The output prediction probability can be expressed as $y_t \sim \text{softmax}(q_t)$.

\subsection{Optimization via Policy Gradient}\label{Optimization}
\vspace{-0.5em}
We formulate the coordination of perturbation selection and saliency fusion as a Reinforcement Learning (RL) \cite{sutton1998reinforcement, li2017deep} problem to navigate the non-differentiable nature of discrete tool selection. By interacting with the frozen VLLM, the APC and SGE learn to maximize the suppression of hallucination pathways.

\vspace{-0.8em}
\paragraph{Policy.}
To facilitate the generation of negative samples, the APC is modeled via independent Bernoulli distributions, allowing for the simultaneous sampling of multiple perturbation tools. Conversely, the SGE fusion weight $\beta$ is treated as a continuous action sampled from a Normal distribution $\mathcal{N}(\mu, \sigma^2)$ during training to encourage exploration.

\vspace{-0.8em}
\paragraph{Reward and Policy Update.}
To guide the learning of both APC and SGE, we define a sparse reward $R$ based on the VLLM's final decision accuracy. The reward is determined by comparing the extracted prediction (\eg, “Yes/No” or “A/B/C”) with the ground truth:
\begin{equation}
    R = \begin{cases} +1.0 & \text{if } \text{prediction} = \text{ground truth}, \\ 
    -1.0 & \text{otherwise}. \end{cases}
\end{equation}
We employ the REINFORCE algorithm \cite{williams1992simple, sutton1998reinforcement} with a moving baseline $B$ to update the policy parameters. The baseline is updated via an exponential moving average ($B \leftarrow \varphi \cdot B + (1-\varphi) \cdot R$) to reduce gradient variance. The objective is to minimize the following joint loss:
\begin{equation}
\mathcal{L} = - (\log P(\mathbf{a}_{apc} \mid \mathbf{H},\mathbf{D}) + \log P(\mathbf{a}_{sge} \mid \mathbf{H})) \cdot (R - B)  
\end{equation} 
where $\mathbf{a}_{apc}$ and $\mathbf{a}_{sge}$ represent the sampled perturbation mask and fusion weight, respectively. Further details are provided in \sec \ref{appendix_opt}.

\begin{table}[!t]
  \centering
  \caption{Benchmark results on Accuracy. The shaded cells highlight the \colorbox{best}{best} and \colorbox{second}{second-best} results in the open-source and commercial models, respectively. The \textit{\textbf{best model result}} among all models is highlighted. All results can be seen in \sec \ref{Results}}
  \vspace{-0.5em}
  \label{tab_benchmark_results}
  \setlength{\tabcolsep}{1.5pt}
  \resizebox{0.47\textwidth}{!}{ 
  \begin{tabular}{lcccc} 
    \toprule
    \textbf{Model} & \textbf{Size} & \textbf{Real} & \textbf{Generated} & \textbf{Avg} \\
    \midrule
Human&-& \textbf{0.95} & \textbf{0.94} & \textbf{0.95} \\
\midrule
\multicolumn{5}{c}{\textit{Open-source Models}} \\
\midrule
VideoChat-Flash \cite{li2024videochatflash} & 2B & 0.55 & 0.57 & 0.56 \\
VideoLLaMA3 \cite{damonlpsg2025videollama3} & 2B & 0.51 & 0.58 & 0.54 \\
Qwen3-VL-Instruct \cite{Qwen3-VL} & 2B & 0.59 & 0.67 & 0.63 \\
Qwen3-VL-Thinking \cite{Qwen3-VL} & 2B & 0.62 & 0.69 & 0.65 \\
Qwen2.5-VL-Instruct \cite{bai2025qwen2_5} & 3B & 0.52 & 0.61 & 0.57 \\
InternVL3.5 \cite{wang2025internvl3_5} & 4B & 0.61 & 0.66 & 0.64 \\
MiniCPM-V-4 \cite{yu2025minicpmv45cookingefficient} & 4B & 0.57 & 0.63 & 0.60 \\
Qwen3-VL-Instruct \cite{Qwen3-VL} & 4B & 0.64 & 0.72 & 0.68 \\
Qwen3-VL-Thinking \cite{Qwen3-VL} & 4B & 0.67 & 0.73 & 0.70 \\
Molmo2 \cite{clark2026molmo2openweightsdata} & 4B & 0.63 & 0.68 & 0.66 \\
LLaVA-NeXT-Video \cite{zhang2024llavanext-video} & 7B & 0.44 & 0.49 & 0.47 \\
VideoChat-Flash \cite{li2024videochatflash} & 7B & 0.58 & 0.60 & 0.59 \\
VideoLLaMA3 \cite{damonlpsg2025videollama3} & 7B & 0.59 & 0.65 & 0.62 \\
Qwen2.5-VL-Instruct \cite{bai2025qwen2_5} & 7B & 0.60 & 0.69 & 0.64 \\
InternVL3.5 \cite{wang2025internvl3_5} & 8B & 0.60 & 0.67 & 0.64 \\
MiniCPM-V-4.5 \cite{yu2025minicpmv45cookingefficient} & 8B & 0.64 & 0.68 & 0.66 \\
Qwen3-VL-Instruct \cite{Qwen3-VL} & 8B & 0.65 & 0.74 & 0.70 \\
Qwen3-VL-Thinking \cite{Qwen3-VL} & 8B & 0.67 & 0.74 & 0.71 \\
Molmo2 \cite{clark2026molmo2openweightsdata} & 8B & 0.65 & 0.69 & 0.67 \\
Kimi-VL-Instruct \cite{kimiteam2025kimivltechnicalreport} & 16B & 0.64 & 0.72 & 0.68 \\
Kimi-VL-Thinking \cite{kimiteam2025kimivltechnicalreport} & 16B & 0.68 & 0.73 & 0.71 \\
InternVL3.5 \cite{wang2025internvl3_5} & 30B & 0.62 & 0.67 & 0.65 \\
Qwen2.5-VL-Instruct \cite{bai2025qwen2_5} & 32B & 0.62 & 0.71 & 0.67 \\
Qwen3-VL-Instruct \cite{Qwen3-VL} & 32B & 0.69 & 0.76 & 0.72 \\
Qwen3-VL-Thinking \cite{Qwen3-VL} & 32B & \cellcolor{second} 0.71 & \cellcolor{second} 0.77 & \cellcolor{second} 0.74 \\
LLaVA-NeXT-Video\cite{zhang2024llavanext-video} & 34B & 0.47 & 0.53 & 0.50 \\
GLM-4.5v \cite{vteam2025glm45vglm41vthinkingversatilemultimodal} & 108B & 0.47 & 0.50 & 0.49 \\
GLM-4.6v-flash \cite{vteam2025glm45vglm41vthinkingversatilemultimodal} & 108B & 0.44 & 0.49 & 0.46 \\
GLM-4.6v \cite{vteam2025glm45vglm41vthinkingversatilemultimodal} & 108B & 0.49 & 0.52 & 0.51 \\
Qwen3-VL-Instruct \cite{Qwen3-VL} & 235B & 0.71 & 0.77 & 0.74 \\
Qwen3-VL-Thinking \cite{Qwen3-VL} & 235B & \cellcolor{best} 0.73 & \cellcolor{best} \textit{\textbf{0.78}} & \cellcolor{best} 0.76 \\
\midrule
\multicolumn{5}{c}{\textit{Proprietary Models}} \\
\midrule
GPT-4o \cite{openai2025gpt4o} & - & 0.70 & 0.73 & 0.71 \\
GPT-5 \cite{openai2025gpt5} & - & 0.72 & 0.75 & 0.73 \\
GPT-5.2 \cite{openai2025gpt5_2} & - & 0.74 & \cellcolor{best} 0.77 & \cellcolor{second} 0.76 \\
Gemini-2.5-flash \cite{google2025gemini2_5flash} & - & 0.71 & 0.75 & 0.73 \\
Gemini-2.5-pro \cite{google2025gemini2_5pro} & - & 0.71 & 0.74 & 0.73 \\
Gemini-3-pro \cite{google2025gemini3pro} & - & \cellcolor{best} \textit{\textbf{0.76}} & \cellcolor{second} 0.77 & \cellcolor{best} \textit{\textbf{0.77}} \\
Doubao-Seed-1.6 \cite{doubao2025seed1_6} & - & 0.71 & 0.75 & 0.73 \\
Doubao-Seed-1.8 \cite{doubao2025seed1_8} & - & \cellcolor{second} 0.73 & 0.77 & 0.75 \\
    \bottomrule
  \end{tabular}
  }
  \vspace{-1.5em}
\end{table}

\section{Experiments}
\vspace{-0.5em}
\subsection{Experimental Settings}
\vspace{-0.5em}
\paragraph{Models}
We evaluate 39 VLLMs spanning 10 open-source and 3 proprietary model families, and include a human baseline for reference. Model details and implementation details are provided in \sec \ref{Models} and \sec \ref{Implementation_Details}, respectively. 

\vspace{-0.8em}
\paragraph{Evaluation Metrics.}
We adopt Accuracy as the primary metric for both tasks, and additionally report task-specific diagnostic metrics, including Yes/No Accuracy for YNQA and macro-averaged Precision, Recall, and F1-score for MCQA. Detailed definitions are provided in \sec \ref{Metrics Details}.

\subsection{Benchmarking Models on \benchmark{}}
\begin{figure}[!tb]
\centering
\includegraphics[width=0.87\linewidth]{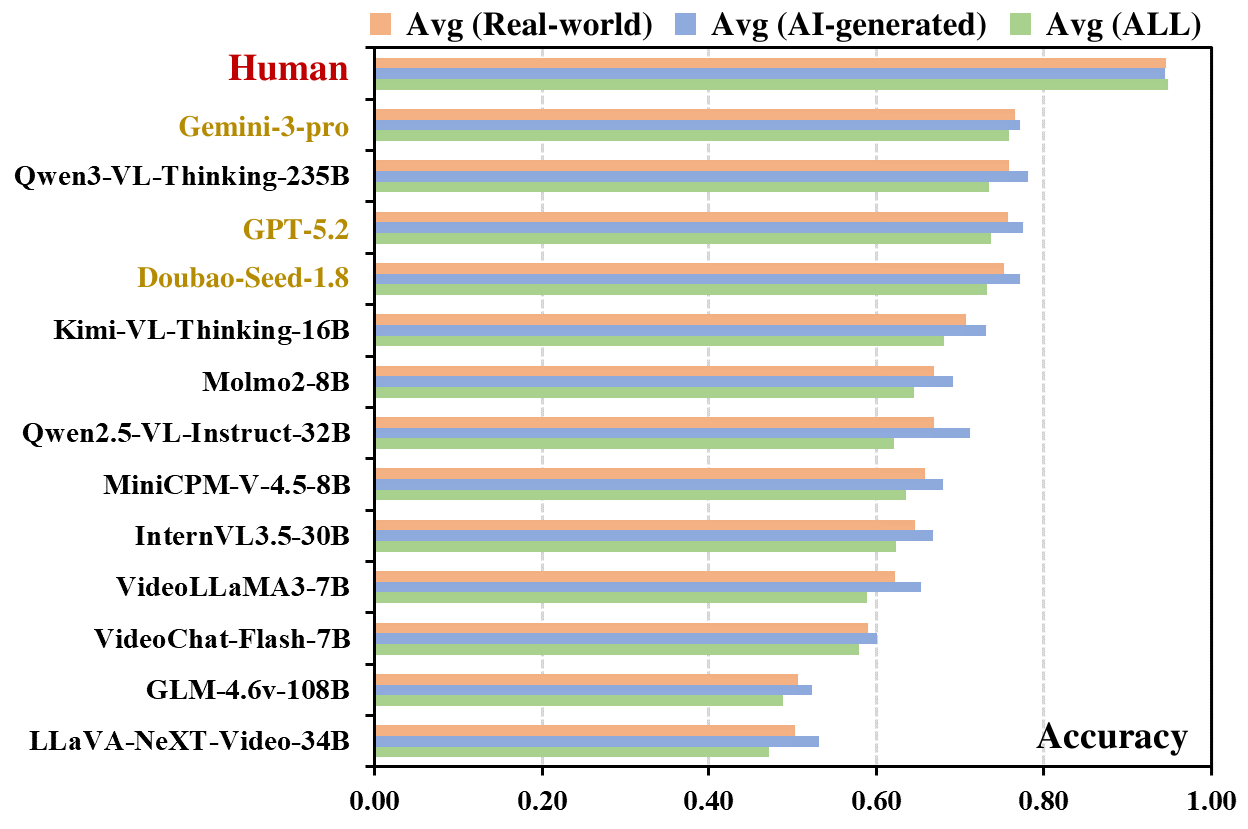}
\vspace{-0.5em}
\caption{Comparative performance of the leading representatives from each model family. Human performance is highlighted at the top to establish a ceiling for evaluating current VLLM capabilities.}
\label{fig_main_results1}
\vspace{-1.0em}
\end{figure}

\vspace{-0.5em}
\tab \ref{tab_benchmark_results} systematically reports the performance of 39 current VLLMs on \benchmark{} on Accuracy. The multi-dimensional analysis reveals the following:

\textbf{Obs.\ding{182} Proprietary models lead in performance, though top-tier open-source models are narrowing the gap; a significant gap persists between AI and stable human performance.} As shown in \fig \ref{fig_main_results1}, proprietary models such as Gemini-3-pro (0.77 avg) and GPT-5.2 (0.76 avg) establish a high performance floor with superior architectural robustness across diverse tasks. Within the open-source community, a stark performance divide exists where smaller models like VideoLLaMA3-2B (0.54 avg) struggle with spatiotemporal grounding, while high-end scaling in Qwen3-VL-Thinking-235B (0.76 avg) effectively achieves parity with proprietary state-of-the-art capabilities. Despite these advancements, a substantial 18\% accuracy gap remains between the best-performing AI and the human baseline (0.95), whose remarkably consistent performance (0.91–0.96) across all complexities serves as critical evidence that \benchmark{} is a well-calibrated diagnostic tool, ensuring that degradation is strictly attributable to inherent reasoning failures rather than dataset bias.

\begin{figure}[!t]
\centering
\includegraphics[width=0.82\linewidth]{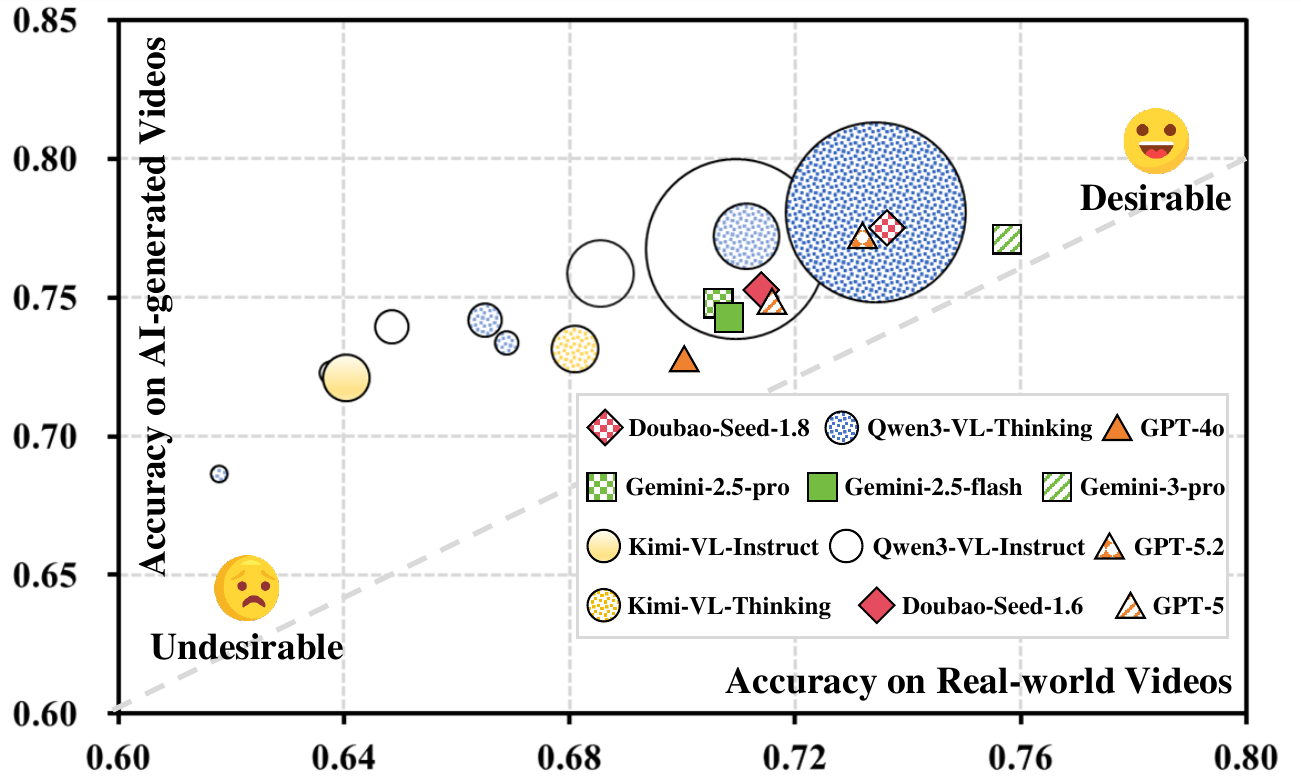}
\vspace{-0.5em}
\caption{Performance correlation between different domains across VLLMs. For open-source models, circle sizes represent parameter scale: the Qwen3-VL series ranges from 2B to 235B, and the Kimi-VL series is 16B. The visualization further distinguishes between the Thinking and Instruct reasoning paradigms.}
\vspace{-1.0em}
\label{fig_main_results2}
\end{figure}

\textbf{Obs.\ding{183} Scale and reasoning paradigms influence performance; real-world videos are more challenging than generated ones—this gap motivated our targeted investigation into the intricate compositional hallucinations.} As shown in \fig \ref{fig_main_results2}, models utilizing the “Thinking” paradigm tend to outperform their “Instruct” counterparts, such as Qwen3-VL-Thinking-2B (0.65 avg) \vs Qwen3-VL-Instruct-2B (0.63 avg). suggesting that internal reasoning processes help mitigate logical biases across modalities. Larger models generally perform better. Proprietary models show consistent improvements over iterations, such as GPT-5 (0.73 avg) outperforming GPT-4o (0.71 avg). The positioning of most models above the diagonal in \fig \ref{fig_main_results2} highlights a significant domain bias: current VLLMs exhibit a higher risk of hallucination in real-world scenarios than in AI-generated content. While AI-generated content is curated to trigger hallucinations through unseen scenarios, our findings reveal that they still lack the complex dynamics and lighting of real-world videos. Crucially, this gap also reveals that the primary bottleneck is not scenario novelty but the dense spatiotemporal entanglement of real-world events, which triggers more elusive compositional hallucinations that are far more difficult to resolve. Based on this, we introduce \benchmark{} as a rigorous diagnostic tool and propose \approach{} to calibrate the model's decoding boundaries against such multi-faceted logical contradictions. 

\begin{figure*}[!t]
\centering
  \begin{minipage}{0.78\linewidth} 
    \centering
    \includegraphics[width=\linewidth]{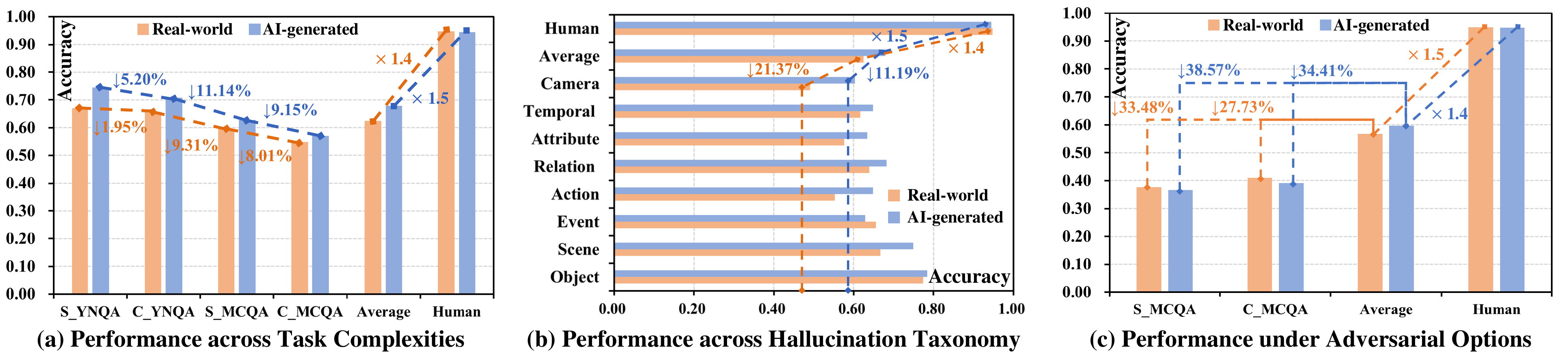}
    \vspace{-1.0em}
    \caption{Performance analysis on \benchmark{}. (a) Accuracies decline from S\_YNQA to C\_MCQA. (b) Performance across eight types reveals that models struggle most with the novel camera type. (c) Significant accuracy drops under adversarial options reveal a severe lack of predictive stability in VLLMs.}
    \label{fig_main_results3}
  \end{minipage}
  \hfill
  \begin{minipage}{0.20\linewidth} 
    \centering
    \includegraphics[width=\linewidth]{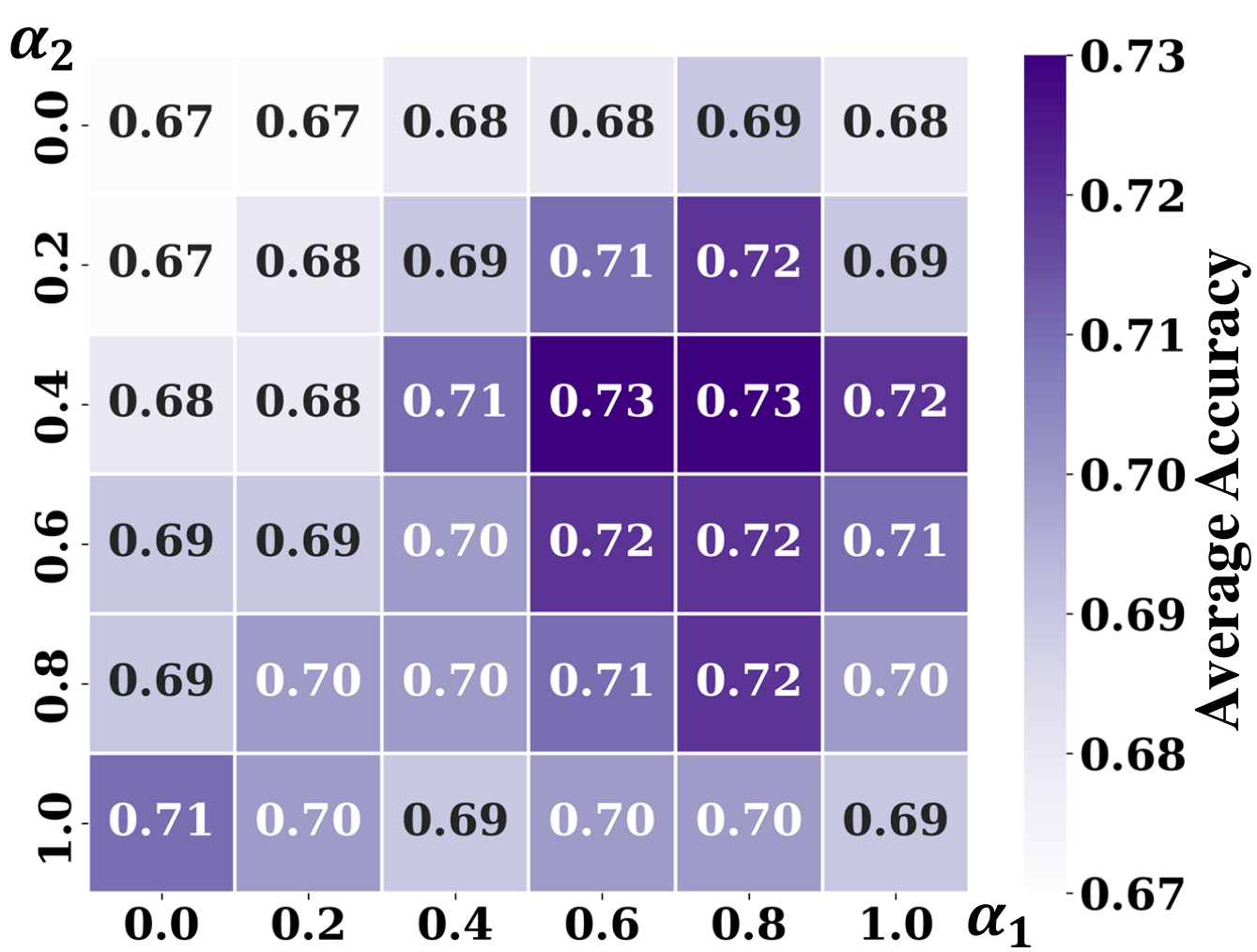}
    \vspace{-1.0em}
    \caption{Hyperparameter sensitivity. The optimal combination is $\alpha_1=0.8$ and $\alpha_2=0.4$.}
    \label{fig_hyper}
  \end{minipage}  
  \vspace{-1.0em}
\end{figure*}

\textbf{Obs.\ding{184} Performance decreases with compositional complexity; camera-based reasoning is a dominant failure mode; adversarial options expose pervasive model instability.} As shown in the left panel of \fig \ref{fig_main_results3}, accuracy drops from S\_YNQA to C\_MCQA, highlighting the need for compositional hallucination detection. The middle panel in \fig \ref{fig_main_results3} indicates that models exhibit the highest error rates when addressing camera-based movements and perspective shifts. This validates our introduction of this new category, as it uncovers a systematic deficiency in perceiving “observer-perspective dynamics”. The right panel shows that adversarial options significantly increase hallucination rates, pointing to higher instability under such conditions. These findings underline the challenges of \benchmark{} and establish a foundation for developing more robust models. Driven by these insights, we specifically engineered \approach{}, incorporating APC to suppress the identified reasoning instabilities and SGE to resolve the pervasive spatiotemporal grounding deficiencies in complex scenarios.

\vspace{-0.8em}
\paragraph{Summary.}
\benchmark{} not only reveals a significant performance gap between current models and humans ($\textgreater 0.15$) but also precisely localizes the weaknesses in spatiotemporal logic, complex reasoning, and camera dynamics. 

\subsection{Main Results of \approach{}}
\vspace{-0.5em}
We evaluate the effectiveness of \approach{} by comparing it with representative contrastive decoding baselines, including MotionCD \cite{kong2025mhbench}, TCD \cite{zhang2024eventhallusion}, and DINO-HEAL \cite{li2025vidhalluc} (as detailed in \sec \ref{Baseline Methods}).

As shown in \tab \ref{tab_ominivcd_results}, \approach{} consistently delivers substantial performance gains across diverse model architectures. On Qwen3-VL-Instruct-8B, it boosts average accuracy from 0.67 to 0.73; on VideoLLaMA3-7B, it improves performance by 12.03\%. These results highlight its robust scalability and plug-and-play compatibility with various open-source VLLMs. Compared to existing baselines, our \approach{} significantly outperforms methods that rely on uniform perturbations or a single saliency map. While MotionCD and TCD provide marginal gains ($\approx 1.8\%-3.6\%$) by targeting temporal inconsistencies. DINO-HEAL shows limited improvement ($\approx 1.9\%-2.9\%$) in dynamic scenarios as it lacks explicit motion anchoring. In contrast, our triple-pathway framework achieves a much higher performance ceiling by concurrently leveraging context-aware negative distraction and spatiotemporal evidence enhancement. We also report the inference efficiency in \sec \ref{Optimization and Efficiency}.

\begin{table}[!b]
\vspace{-1.0em}
\centering
\caption{Main results on the test set. The shaded cells highlight the \colorbox{second}{backbone's} and \colorbox{best}{our method's} results. The final column reports the relative percentage against the original backbone's accuracy.}
\vspace{-0.5em}
\setlength{\tabcolsep}{1pt}
\label{tab_ominivcd_results}
\resizebox{0.48\textwidth}{!}{
\begin{tabular}{lccccccc}
\toprule
\textbf{Model} & \textbf{S\_YNQA} & \textbf{C\_YNQA} & \textbf{S\_MCQA} & \textbf{C\_MCQA} & \textbf{Avg} & - \\ 
\midrule
\rowcolor{second} Qwen3-VL-Instruct-8B & 0.75 & \textbf{\textit{0.69}} & 0.61 & \textbf{\textit{0.53}} & 0.67 & - \\
+MotionCD & 0.76 & \textbf{\textit{0.70}} & 0.62 & \textbf{\textit{0.54}} & 0.68 & $\uparrow$1.82\% \\
+TCD & 0.77 & \textbf{\textit{0.72}} & 0.62 & \textbf{\textit{0.55}} & 0.69 & $\uparrow$3.61\% \\
+DINO-HEAL & 0.77 & \textbf{\textit{0.72}} & 0.62 & \textbf{\textit{0.54}} & 0.68 & $\uparrow$1.99\% \\
\rowcolor{best} +\approach{} & 0.79 & \textbf{\textit{0.78}} & 0.66 & \textbf{\textit{0.64}} & 0.73 & $\uparrow$9.61\% \\
\midrule
\rowcolor{second} VideoLLaMA3-7B & 0.67 & \textbf{\textit{0.61}} & 0.56 & \textbf{\textit{0.48}} & 0.60 & - \\
+MotionCD & 0.68 & \textbf{\textit{0.62}} & 0.57 & \textbf{\textit{0.48}} & 0.61 & $\uparrow$2.34\% \\
+TCD & 0.69 & \textbf{\textit{0.63}} & 0.57 & \textbf{\textit{0.49}} & 0.62 & $\uparrow$3.41\% \\
+DINO-HEAL & 0.69 & \textbf{\textit{0.63}} & 0.57 & \textbf{\textit{0.48}} & 0.62 & $\uparrow$2.92\% \\
\rowcolor{best} +\approach{} & 0.74 & \textbf{\textit{0.69}} & 0.63 & \textbf{\textit{0.55}} & 0.67 & $\uparrow$12.03\% \\
\bottomrule
\end{tabular}
}
\end{table}

\begin{table}[!b]
\centering
\caption{Ablation study of \approach{}. The shaded cells highlight the \colorbox{second}{backbone's} and \colorbox{best}{our method's} results. The final column reports the relative percentage against the original backbone's accuracy.}
\vspace{-0.5em}
\setlength{\tabcolsep}{1pt}
\label{tab_ablation}
\resizebox{0.48\textwidth}{!}{
\begin{tabular}{lcccccc}
\toprule
\textbf{Model} & \textbf{S\_YNQA} & \textbf{C\_YNQA} & \textbf{S\_MCQA} & \textbf{C\_MCQA} & \textbf{Avg} & -\\ 
\midrule
\rowcolor{second} Qwen3-VL-Instruct-8B & 0.75 & \textbf{\textit{0.69}} & 0.61 & \textbf{\textit{0.53}} & 0.67 & - \\
\rowcolor{best} +\approach{} & 0.79 & \textbf{\textit{0.78}} & 0.66 & \textbf{\textit{0.64}} & 0.73 & $\uparrow$9.61\% \\
\midrule
+\approach{} w/o APC & 0.78 & \textbf{\textit{0.72}} & 0.62 & \textbf{\textit{0.54}} & 0.69 & $\uparrow$2.54\% \\
+\approach{} w/o SGE & 0.78 & \textbf{\textit{0.76}} & 0.63 & \textbf{\textit{0.61}} & 0.71 & $\uparrow$6.72\% \\
+\approach{} w/ RP & 0.70 & \textbf{\textit{0.69}} & 0.54 & \textbf{\textit{0.52}} & 0.63 & $\downarrow$5.87\% \\
+\approach{} w/o DINO & 0.78 & \textbf{\textit{0.77}} & 0.64 & \textbf{\textit{0.63}} & 0.72 & $\uparrow$7.74\% \\
+\approach{} w/o Motion & 0.79 & \textbf{\textit{0.77}} & 0.64 & \textbf{\textit{0.62}} & 0.72 & $\uparrow$7.65\% \\
\bottomrule
\end{tabular}
}
\end{table}

A critical observation from the results is that our \approach{} provides the most substantial improvements in compositional (C\_) tasks. For instance, Qwen3-VL-Instruct-8B sees a 9\% jump in C\_YNQA and an 11\% increase in C\_MCQA. This performance leap is a direct result of the synergistic dual action of the APC and SGE modules, which together provide a robust “upper and lower bound” calibration for the model's predictions. While the SGE module establishes a foundation of positive anchoring by reinforcing critical spatiotemporal visual evidence, the APC module contributes dynamic negative suppression by filtering out context-irrelevant hallucinations through adaptive perturbations. This dual-boundary calibration effectively bridges the gap in spatiotemporal reasoning that single-pathway baselines fail to address.

\subsection{Ablation Study}
\vspace{-0.5em}

As shown in \tab \ref{tab_ablation}, we conduct a series of ablation studies to evaluate the contribution of each component. Notably, removing APC leads to a more pronounced performance drop (0.73 $\rightarrow$ 0.69) compared to ablating SGE (0.73 $\rightarrow$ 0.71). This degradation is especially evident in compositional hallucination tasks: without APC, accuracy plummets by 0.06 on C\_YNQA and by 0.10 on C\_MCQA. This suggests that adaptive negative suppression plays a more dominant role in resolving hallucinations. Using random perturbation (RP) results in a collapse to 0.63, worse than the backbone (0.67), proving that APC's optimized selection is vital to avoid destructive interference during decoding.  

Within the SGE module, ablating DINOv3 spatial or motion-based temporal features reduces performance to 0.72. The specific drop in C\_MCQA without motion cues (0.64 $\rightarrow$ 0.62) proves the necessity of inter-frame dynamics for complex temporal reasoning.

Finally, sensitivity analysis in \fig \ref{fig_hyper} confirms that the optimal balance is achieved at $\alpha_1 = 0.8$ and $\alpha_2 = 0.4$ (as detailed in \sec \ref{Hyperparameter Analysis}). This peak validates that precisely calibrating the relative strengths of negative distraction and positive evidence is critical for robust spatiotemporal grounding.

\section{Conclusion}
\vspace{-0.5em}
In this paper, we introduce the first comprehensive benchmark (\textbf{\benchmark}) for evaluating both isolated and compositional hallucinations in video multimodal large language models. We also develop a triple-pathway contrastive decoding framework (\textbf{\approach}) that unifies adaptive negative suppression with saliency-guided enhancement for mitigating \textit{compositional} hallucinations. We hope this work will inspire future research on principled modeling and mitigation of compositional hallucinations, towards more robust spatiotemporal reasoning and grounding mechanisms.

\section*{Impact Statement}
This paper introduces the \benchmark{} benchmark and the \approach{} framework to systematically diagnose and mitigate compositional hallucinations in video multimodal large language models. By improving the spatiotemporal grounding and logical consistency of these models, our work contributes to the development of more reliable and trustworthy AI systems. 
The potential societal implications are multifaceted: enhancing the fidelity of video understanding can reduce the generation of misleading or factually incorrect content in automated video analysis, content moderation, and human-AI interaction.

\bibliography{reference}
\bibliographystyle{icml2026}

\newpage
\appendix
\onecolumn

\vspace*{0.5cm}
\begin{center}
    \LARGE \textbf{Supplementary Material for:} \\
    \vspace{0.5em}
    \Large \textit{Learning to Decode Against Compositional Hallucination in \\ Video Multimodal Large Language Models}
\end{center}
\vspace{1cm}
\section*{Appendix Overview}
\addcontentsline{toc}{section}{Appendix Overview}

\begin{center}
{ 
\setlength{\tabcolsep}{4pt} 
\renewcommand{\arraystretch}{1.3}
\begin{tabular}{@{}p{0.85\linewidth}r@{}}

\hyperref[Related Work]{\textbf{A. Related Work}} \dotfill & \textbf{\pageref{Related Work}} \\
\hspace{1.5em}\hyperref[Video Multimodal Large Language Models]{Video Multimodal Large Language Models} \dotfill & \pageref{Video Multimodal Large Language Models} \\
\hspace{1.5em}\hyperref[Video Hallucination Benchmark]{Video Hallucination Benchmark} \dotfill & \pageref{Video Hallucination Benchmark} \\
\hspace{1.5em}\hyperref[Hallucination Mitigation Approach]{Hallucination Mitigation Approach} \dotfill & \pageref{Hallucination Mitigation Approach} \\

\hyperref[Benchmark Details]{\textbf{B. OmniVCHall Benchmark Details}} \dotfill & \textbf{\pageref{Benchmark Details}} \\
\hspace{1.5em}\hyperref[Hallucination Type]{Hallucination Type} \dotfill & \pageref{Hallucination Type} \\
\hspace{1.5em}\hyperref[Data Sources]{Data Source} \dotfill & \pageref{Data Sources} \\
\hspace{1.5em}\hyperref[Prompt for QA Generation]{Prompt for QA Generation} \dotfill & \pageref{Prompt for QA Generation} \\
\hspace{1.5em}\hyperref[Adversarial Option]{Adversarial Option} \dotfill & \pageref{Adversarial Option} \\
\hspace{1.5em}\hyperref[QA Distribution]{QA Distribution} \dotfill & \pageref{QA Distribution} \\
\hspace{1.5em}\hyperref[Word Cloud]{Word Cloud} \dotfill & \pageref{Word Cloud} \\
\hspace{1.5em}\hyperref[sec:iaa]{Inter-Annotator Agreement (IAA)} \dotfill & \pageref{sec:iaa} \\

\hyperref[Framework Details]{\textbf{C. TriCD Framework Details}} \dotfill & \textbf{\pageref{Framework Details}} \\
\hspace{1.5em}\hyperref[Tool Analysis]{Tool Analysis} \dotfill & \pageref{Tool Analysis} \\
\hspace{1.5em}\hyperref[Visual Examples]{Visual Examples} \dotfill & \pageref{Visual Examples} \\
\hspace{1.5em}\hyperref[appendix_apc]{Details of Adaptive Perturbation Controller} \dotfill & \pageref{appendix_apc} \\
\hspace{1.5em}\hyperref[appendix_sge]{Details of Saliency-Guided Enhancement} \dotfill & \pageref{appendix_sge} \\
\hspace{1.5em}\hyperref[appendix_opt]{Details of Optimization} \dotfill & \pageref{appendix_opt} \\

\hyperref[Experimental Details]{\textbf{D. Experimental Details}} \dotfill & \textbf{\pageref{Experimental Details}} \\
\hspace{1.5em}\hyperref[Models]{Model Details} \dotfill & \pageref{Models} \\
\hspace{1.5em}\hyperref[Metrics Details]{Metrics Details} \dotfill & \pageref{Metrics Details} \\
\hspace{1.5em}\hyperref[Implementation_Details]{Implementation Details} \dotfill & \pageref{Implementation_Details} \\
\hspace{1.5em}\hyperref[Results]{Benchmark Results} \dotfill & \pageref{Results} \\
\hspace{1.5em}\hyperref[Baseline Methods]{Baseline Methods} \dotfill & \pageref{Baseline Methods} \\
\hspace{1.5em}\hyperref[Optimization and Efficiency]{Optimization and Efficiency} \dotfill & \pageref{Optimization and Efficiency} \\
\hspace{1.5em}\hyperref[Hyperparameter Analysis]{Hyperparameter Analysis} \dotfill & \pageref{Hyperparameter Analysis} \\
\hspace{1.5em}\hyperref[Case Study]{Case Study} \dotfill & \pageref{Case Study} \\
\end{tabular}
}
\end{center}

\newpage

\section{Related Work} \label{Related Work}
\subsection{Video Multimodal Large Language Models} \label{Video Multimodal Large Language Models}
Multimodal large language models have achieved remarkable success in image understanding \cite{liu2023visual, achiam2023gpt} by effectively aligning visual features with linguistic representations in a shared latent space \cite{radford2021learning, li2023blip}. Building on this foundation, the field has increasingly pivoted toward video multimodal large language models (VLLMs) to model complex temporal dynamics across high-dimensional visual sequences \cite{zou2024seconds, tang2025video}. Most recently, frontier closed source models such as GPT-5.2 \cite{openai2025gpt5_2} and Gemini-3-Pro \cite{google2025gemini3pro} have emerged, as well as high-performance open source models including GLM-4.6V \cite{vteam2025glm45vglm41vthinkingversatilemultimodal}, Qwen3-VL \cite{Qwen3-VL}, InternVL3.5 \cite{wang2025internvl3_5}, and VideoLLaMA3 \cite{damonlpsg2025videollama3}. By incorporating sophisticated spatiotemporal compression techniques such as token reduction \cite{li2024llama}, merging \cite{jin2024chat}, or pooling \cite{xu2024pllava}, VLLMs have achieved unprecedented performance in perception and efficiency. However, current VLLMs remain profoundly susceptible to severe hallucination issues, which continue to pose a major obstacle for their deployment in reliability-critical scenarios \cite{rawte2023survey, sahoo2024comprehensive}.

\subsection{Video Hallucination Benchmark} \label{Video Hallucination Benchmark}
The rapid evolution of VLLMs has catalyzed the development of various evaluation benchmarks such as SEED-Bench \cite{li2023seed}, MVBench \cite{li2024mvbench}, and VideoMME \cite{fu2025video}. And the quantification of visual hallucinations has become a pivotal necessity for ensuring reliability. Consequently, specialized benchmarks including VidHal \cite{choong2024vidhal}, VideoCon \cite{bansal2024videocon}, VideoHallucer \cite{wang2024videohallucer}, EventHallusion \cite{zhang2024eventhallusion}, MHBench \cite{kong2025mhbench}, VidHalluc \cite{li2025vidhalluc}, and ARGUS \cite{rawal2025argus} have been introduced to detect specific visual inconsistencies. Despite these advancements, the evaluative scope of existing benchmarks remains notably fragmented. Most current efforts focus on isolated hallucination types such as action-based \cite{kong2025mhbench} or temporal-based \cite{choong2024vidhal, li2025vidhalluc} while neglecting essential video-centric elements, particularly camera-lens dynamics. Furthermore, these works often overlook the complexity of compositional hallucinations \cite{yang2025discovering, chytas2025reco}, where multiple hallucination types concurrently manifest within a single query. These limitations underscore the need for a more comprehensive and integrated evaluation protocol. As summarized in \tab \ref{tab_other_datasets}, our \benchmark{} fills this gap by introducing eight fine-grained hallucination types and complex multi-type testing to assess the robustness of VLLMs across both real-world and generated videos.

\subsection{Hallucination Mitigation Approach}  \label{Hallucination Mitigation Approach}
Hallucination manifests as generated responses that contain information inconsistent with the actual visual content, typically arising from modal misalignment, imbalanced training data, or over-reliance on prior linguistic knowledge \cite{bansal2024videocon, bai2024hallucination}. Various strategies have been proposed to mitigate these inconsistencies. For instance, a contrast-caption fine-tuning strategy \cite{bansal2024videocon} improves video-language correspondence. The Self-PEP framework \cite{wang2024videohallucer} utilizes a Predict-Explain-Predict loop to verify and revise initial judgments through generated explanations. DINO-HEAL \cite{li2025vidhalluc} addresses the vulnerability of VLLMs to semantically similar distractors, attributing such hallucinations to the inherent inductive bias of CLIP-series encoders \cite{radford2021learning, zhai2023sigmoid}. Furthermore, a method employing improved positional encoding and Direct Preference Optimization \cite{lu2025elv} targets semantic aggregation in long-video understanding. Additionally, a hierarchical multimodal consistency approach reduces hallucinations by enforcing structural alignment across multiple granularities of video and language through specialized training objectives \cite{dang2025hallucination}.

Among these advancements, Contrastive Decoding (CD) has emerged as a powerful training-free alternative that refines token predictions by adjusting logit streams during inference \cite{leng2024mitigating, park2025second}. A temporal CD approach \cite{zhang2024eventhallusion} has been developed by contrasting original frames with temporally distorted sequences. Similarly, MotionCD \cite{kong2025mhbench} penalizes logits generated from motion-corrupted negative samples. However, these methods are often restricted to predefined hallucination types and employ static, uniform perturbations, which lack the flexibility to adapt to diverse video content. To bridge these gaps, our \approach{} framework unifies these two paradigms. It dynamically selects perturbation tools according to video and query context for adaptive negative suppression, while fusing spatial and temporal saliency for visual enhancement.

\clearpage

\section{\benchmark{} Benchmark Details}\label{Benchmark Details}
\subsection{Hallucination Type}\label{Hallucination Type}
\tab \ref{tab_definition_and_examples} presents a formal taxonomy of the eight hallucination types featured in \benchmark{}, providing precise definitions and representative examples for each. This systematic categorization spans fundamental object properties and complex spatiotemporal dynamics, including our novel camera-centric type to isolate lens effects from physical motion. By establishing clear ground-truth boundaries, this table serves as a diagnostic reference for identifying specific perceptual and reasoning failures in VLLMs.

\begin{table*}[htbp]
  \centering
  \caption{Definition and examples of the eight hallucination types in \benchmark{}.}
  \label{tab_definition_and_examples}
  \vspace{-0.5em}
  \resizebox{0.90\textwidth}{!}{
  \begin{tabular}{cll} 
    \toprule
    \textbf{Type} & \textbf{Definition} & \textbf{Example} \\
    \midrule
    Object & Presence or identity of physical entities. & Describing a “cat” when only a “dog” exists. \\
    Scene  & The overarching environment or setting. & Misidentifying a “hospital” as a “school”. \\
    Event & High-level semantic units or causal occurrences. & Misinterpreting a “goal scored” as a “missed shot”. \\
    Action & Physical movement or behavioral patterns. & Claiming a person is “running” while they are “walking”. \\
    Relation  & Spatial or logical interactions between entities. & Stating a bowl is “on the table” when it is “under” it. \\
    Attribute  & Static properties like color, size, or material. & Describing a “red ball” as “blue”.\\
    Temporal  & Chronological order or duration of occurrences. & Reversing the order of “pouring water” and “drinking”. \\
    Camera & Perception of cinematic lens dynamics. & Interpreting a “zoom-in” as an “object moving closer”. \\
    \bottomrule
  \end{tabular}
  }
\end{table*}

\subsection{Data Source}\label{Data Sources}
\tab \ref{tab_source} details the distribution of the 823 video clips across 11 heterogeneous sources. The dataset is balanced between established real-world benchmarks and synthetic content from frontier generative models like Sora \cite{openai2025sora}, Veo \cite{google2025veo}, and Wan \cite{alibaba2025wan}. This multi-domain composition ensures that \benchmark{} provides a robust evaluation environment covering both authentic real-world dynamics and complex AI-generated artifacts.

\begin{table}[htbp]
  \centering
  \caption{Statistics of video data sources in \benchmark{}.}
  \label{tab_source}
  \resizebox{0.98\textwidth}{!}{ 
  \begin{tabular}{clcl} 
    \toprule
    \textbf{Type} & \textbf{Source} & \textbf{Count} & \textbf{Platform}\\
    \midrule
    \multirow{4}{*}{Real-world} & VidHalluc \cite{li2025vidhalluc} & 51 & \url{https://huggingface.co/datasets/chaoyuli/VidHalluc} \\
     & Online Search & 100 & \url{https://www.youtube.com/} \\
     & MHBench \cite{kong2025mhbench} & 120 & \url{https://drive.google.com/drive/folders/1INrzOafJe6uKFp0IZp1z-pdw9bq_YYpJ} \\
     & EventHallusion \cite{zhang2024eventhallusion} & 152 & \url{https://drive.google.com/file/d/1IPmx6Y80UrXwVPmZJh6zjCPHtlsw4p9n/view} \\
    \midrule
    \multirow{7}{*}{AI-generated} & Sora \cite{openai2025sora} & 34 & \url{https://openai.com/sora/} \\
     & SeedanceVideo \cite{bytedance2025seedance} & 36 & \url{https://www.seedance.ai/} \\
     & Veo \cite{google2025veo} & 55 & \url{https://deepmind.google/models/veo/} \\
     & HailuoVideo \cite{minimax2025hailuo} & 55 & \url{https://hailuoai.video/} \\
     & HunyuanVideo \cite{tencent2025hunyuan} & 70 & \url{https://www.hunyuanvideo.org/} \\
     & KlingVideo \cite{kuaishou2025kling} & 70 & \url{https://app.klingai.com/global/} \\
     & Wan \cite{alibaba2025wan} & 80 & \url{https://wan.video/} \\
    \bottomrule
  \end{tabular}
  }
\end{table}

\subsection{Prompt for QA Generation}\label{Prompt for QA Generation}
\begin{figure*}[htbp]
\centering
\includegraphics[width=0.98\linewidth]{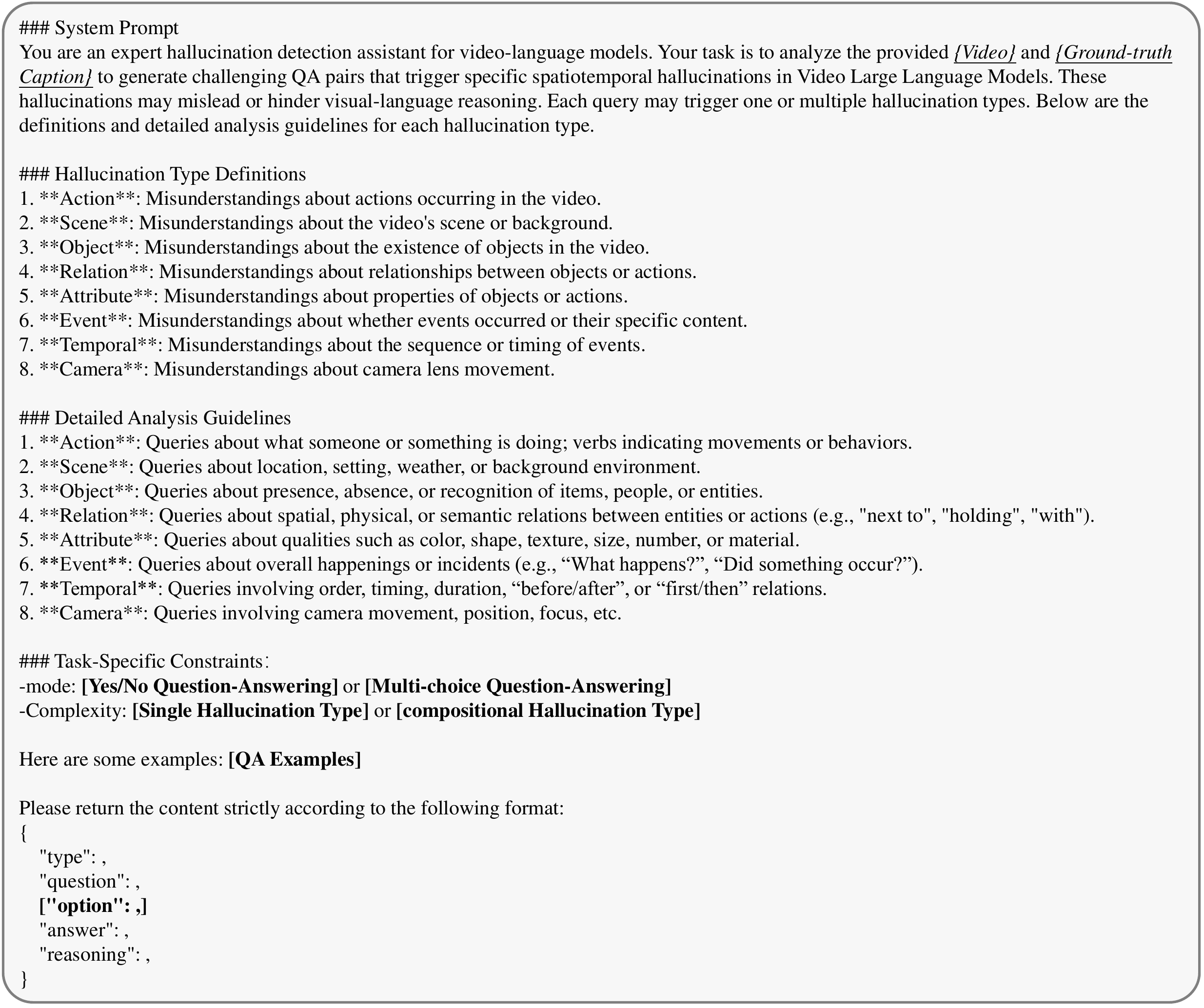}
\vspace{-0.5em}
\caption{Prompt for Gemini-2.5-Pro \cite{google2025gemini2_5pro} to generate QAs. \underline{\textit{Terms}} (\eg, \underline{\textit{\{Video\}}} and \underline{\textit{\{Ground-truth Caption\}}}) represent the multi-modal input variables provided to the model. \textbf{Terms} (\eg, \textbf{[Yes/No Question-Answering]} or \textbf{[Single Hallucination Type]}) denote dynamic slots that are selectively toggled based on the target question format (YNQA \vs{} MCQA) and complexity level (Simple \vs{} Compositional) required for the four evaluation sub-categories.}
\label{fig_qa_prompt}
\end{figure*}

\fig \ref{fig_qa_prompt} illustrates the structured prompt utilized for the automated generation of QA pairs. By providing multi-modal inputs, including the video and its corresponding ground-truth caption, the prompt employs dynamic slots to toggle between different question formats and complexity levels. This systematic approach ensures that the generated 9,027 QA pairs are accurately aligned with our fine-grained hallucination taxonomy while maintaining rigorous logical consistency across both isolated and compositional scenarios.

\clearpage

\subsection{Adversarial Option}\label{Adversarial Option}
\begin{figure*}[h]
\centering
\includegraphics[width=0.95\linewidth]{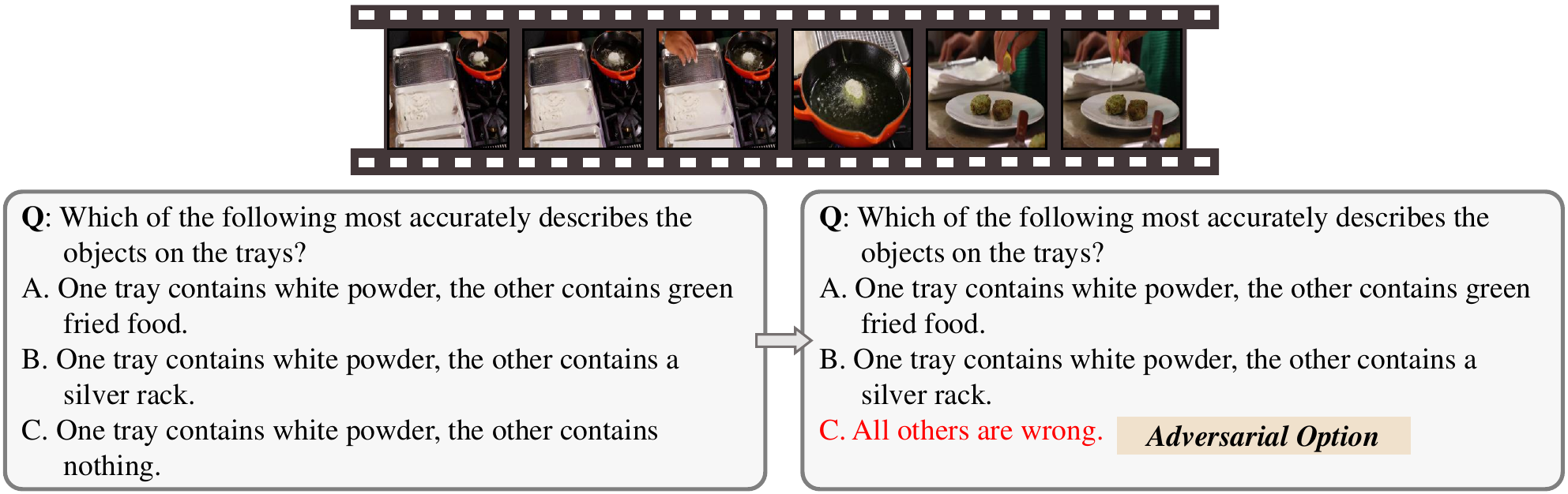}
\vspace{-0.5em}
\caption{Illustration of the “All others are wrong” adversarial injection. (Left) The original MCQA includes a standard set of distractors and a grounded correct answer. (Right) The grounded answer is replaced with the adversarial option “All others are wrong”, requiring the model to explicitly negate all other incorrect descriptions based on the visual evidence.}
\label{fig_ad_example1}
\end{figure*}

\begin{figure*}[h]
\centering
\includegraphics[width=0.95\linewidth]{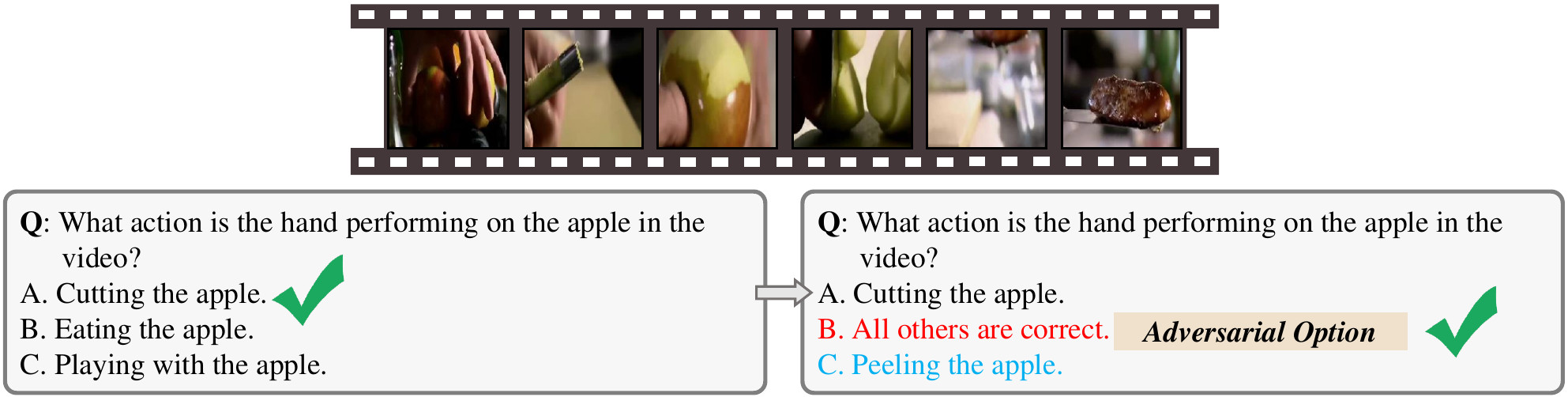}
\vspace{-0.5em}
\caption{Illustration of the “All others are correct” adversarial injection. (Left) A standard MCQA with one correct action and two distractors. (Right) A modified QA where the original error option C is corrected to match the visual content, and option B is subsequently replaced with the adversarial option “All others are correct”, testing the model's ability to recognize multiple valid descriptions simultaneously.}
\label{ad_example2}
\end{figure*}

\tab \ref{tab:omnivh_adversarial} summarizes the distribution of adversarial option injections across different video types and question complexities in \benchmark{}. To enhance evaluation rigor, we selectively replaced standard distractors with adversarial options (\eg, “All others are correct” and “All others are wrong”), maintaining a significant ratio (up to 33.84\% in complex real-world scenarios) to challenge the models' predictive stability, as illustrated in \fig \ref{fig_ad_example1} and \fig \ref{ad_example2}. This injection strategy forces the model to verify every candidate option against the visual evidence rather than relying on a process of elimination or linguistic bias.

\subsection{QA Distribution}\label{QA Distribution}
The detailed distribution of queries across specific hallucination types is provided in \tab \ref{tab_type1} and \tab \ref{tab_type2}. Across both real-world and AI-generated domains, action emerges as the most frequently targeted hallucination type, while all eight categories—including the newly introduced camera type—are extensively covered across simple and complex QA formats.

\begin{table}[htbp]
  \centering
  \caption{Distribution of QA pairs across different hallucination types in real-world scenarios.}
  \label{tab_type1}
  \resizebox{0.75\textwidth}{!}{ 
  \begin{tabular}{cccccccccc} 
    \toprule
    \textbf{Type} & \textbf{Object} & \textbf{Scene} & \textbf{Event} & \textbf{Action} & \textbf{Relation} & \textbf{Attribute} & \textbf{Temporal} & \textbf{Camera} \\
    \midrule
    S\_YNQA & 160 & 99 & 148 & 360 & 206 & 149 & 140 & 153 \\
    C\_YNQA & 467 & 294 & 464 & 525 & 414 & 470 & 306 & 202 \\
    S\_MCQA & 138 & 117 & 93 & 389 & 117 & 82 & 72 & 126 \\
    C\_MCQA & 378 & 260 & 256 & 434 & 267 & 440 & 106 & 91 \\
    \bottomrule
  \end{tabular}
  }
\end{table}

\begin{table}[htbp]
  \centering
  \caption{Distribution of QA pairs across different hallucination types in AI-generated scenarios.}
  \label{tab_type2}
  \resizebox{0.75\textwidth}{!}{ 
  \begin{tabular}{cccccccccc} 
    \toprule
    \textbf{Type} & \textbf{Object} & \textbf{Scene} & \textbf{Event} & \textbf{Action} & \textbf{Relation} & \textbf{Attribute} & \textbf{Temporal} & \textbf{Camera} \\
    \midrule
    S\_YNQA & 254 & 187 & 174 & 330 & 272 & 224 & 199 & 173 \\
    C\_YNQA & 414 & 453 & 422 & 480 & 413 & 553 & 265 & 155 \\
    S\_MCQA & 123 & 151 & 62 & 284 & 146 & 83 & 91 & 172 \\
    C\_MCQA & 330 & 390 & 336 & 399 & 330 & 458 & 246 & 160 \\
    \bottomrule
  \end{tabular}
  }
\end{table}

\clearpage

\subsection{Word Cloud}\label{Word Cloud}
\begin{figure*}[h]
\centering
\includegraphics[width=0.85\linewidth]{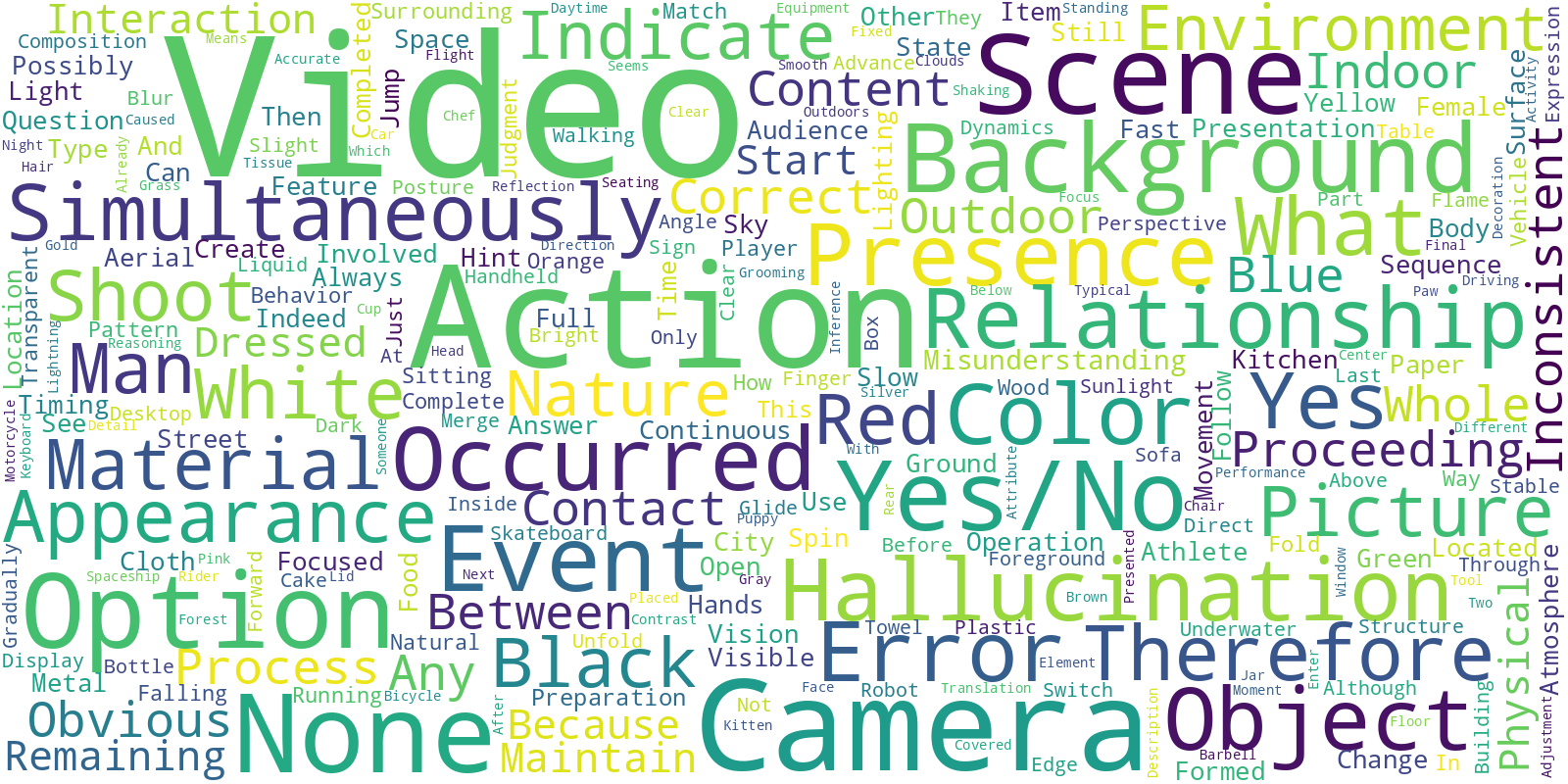}
\vspace{-0.5em}
\caption{Illustration of a semantic word cloud.}
\label{wordcloud_output}
\end{figure*}

As shown in \fig \ref{wordcloud_output}, the word cloud highlights the core focus of our benchmark, particularly emphasizing camera and temporal dynamics.

\subsection{Inter-Annotator Agreement (IAA)}\label{sec:iaa}
To ensure the reliability of the \benchmark{} ground truth, we conducted an IAA analysis across three critical stages:
\begin{itemize}
    \item To validate the quality of the initial video descriptions, we randomly selected 15\% of the video corpus for dual-annotation. Two independent annotators were tasked with identifying primary entities, key actions, and camera movements for each clip. We evaluated their agreement based on the presence and identity of these core spatiotemporal elements. This stage achieved a Cohen’s Kappa of 0.84, representing “almost perfect” agreement according to standard Landis-Koch benchmarks \cite{landis1977measurement}.
    \item The automated QA generation process utilized Gemini-2.5-Pro \cite{google2025gemini2_5pro}. To assess the validity, two independent experts reviewed a stratified random sample of 1,354 pairs (15\%). Each pair was evaluated against three criteria: (1) visual groundedness, (2) logical consistency of the options, and (3) accurate assignment, all with expertise in computer vision, yielding a Cohen’s Kappa of 0.81 for the “Valid/Invalid” classification \cite{landis1977measurement}.
    \item The final reliability of the \benchmark{} ground truth is established through a rigorous human baseline evaluation. We recruited three independent annotators, all with expertise in computer vision, to complete the full evaluation suite. For each of the four evaluation sub-tasks (S\_YNQA, C\_YNQA, S\_MCQA, C\_MCQA), we observed that human accuracy remained remarkably consistent, with performance narrowly constrained between 0.91 and 0.96 regardless of the video source or question complexity. This low variance across sub-tasks demonstrates that the benchmark presents a uniform and fair difficulty level from a human cognitive perspective.
\end{itemize}

\section{\approach{} Framework Details}\label{Framework Details}
\subsection{Tool Analysis}\label{Tool Analysis}
For each tool in $\mathcal{P}$, we construct the comprehensive functional analysis, as summarized in \tab \ref{tab_tools}. These descriptions are formulated using a unified template as presented below:

\begin{quote}
\textit{“This \{Type\} perturbation, specifically \{Description\}, which effectively \{Explain\}”}.
\end{quote}

\begin{table*}[!t]
  \centering
  \caption{Functional analysis of eight video perturbation tools.}
  \label{tab_tools}
  \resizebox{0.98\textwidth}{!}{ 
  \begin{tabular}{cp{1.5cm}p{7.5cm}p{7.5cm}} 
    \toprule
    \textbf{Type} & \textbf{Tool} & \textbf{Description} & \textbf{Explanation} \\
    \midrule
    \multirow{9}{*}{Temporal} & \multirow{3}{*}{Sample} & Performs uniform downsampling at a specific ratio to compress and omit intermediate continuous dynamic transitions. & Disrupts action continuity, forcing the model to reason with sparse temporal evidence. \\
    \cmidrule(lr){2-4}
    & \multirow{3}{*}{Reverse} & Fully reverses the frame sequence chronologically, inverting the causal order while preserving individual frame content. & Penalizes models that rely on static cues rather than true temporal progression (\eg, misidentifying “sitting” vs. “standing”). \\
    \cmidrule(lr){2-4}
    & \multirow{3}{*}{Shuffle} & Randomly permutes the frame order to thoroughly destroy local temporal consistency. & Breaks the logical event chain, exposing models that over-rely on global scene context rather than sequential coherence.  \\
    \midrule
    \multirow{15}{*}{Spatial} & \multirow{3}{*}{Blur} & Applies a Gaussian kernel to systematically weaken high-frequency textures, edges, and fine-grained local details. & Challenges the model’s ability to recognize object categories and existence when fine-grained visual evidence is degraded. \\
    \cmidrule(lr){2-4}
    & \multirow{3}{*}{Noise} & Injects stochastic Gaussian noise to perturb pixel-level consistency while maintaining the global structure. & Exposes vulnerabilities in identifying textures, materials, and surface states that rely on pixel-level statistical accuracy.  \\
    \cmidrule(lr){2-4}
    & \multirow{3}{*}{Grayscale} & Strips all chromatic information by converting RGB frames into a unified luminance-based grayscale format. & Exposes reliance on color-based shortcuts for judging scene attributes, time of day, or specific material properties. \\
    \cmidrule(lr){2-4}
    & \multirow{3}{=}{\centering Horizontal Mirror} & Flips frames horizontally, preserving object identities while altering lateral spatial relationships. & Targets errors in judging camera panning and directional spatial relations (\eg, “left-to-right” vs. “right-to-left” motion). \\ 
    \cmidrule(lr){2-4}
    & \multirow{3}{=}{\centering Vertical Mirror} & Flips frames vertically, inverting top-bottom relationships while maintaining visual consistency. & Penalizes misinterpretations of camera angles (\eg, high vs. low-angle shots) and gravity-based spatial priors (\eg, “above” vs. “below”). \\ 
    \bottomrule
  \end{tabular}
  }
\end{table*}

\subsection{Visual Examples}\label{Visual Examples}

\begin{figure*}[h]
\centering
\includegraphics[width=0.85\linewidth]{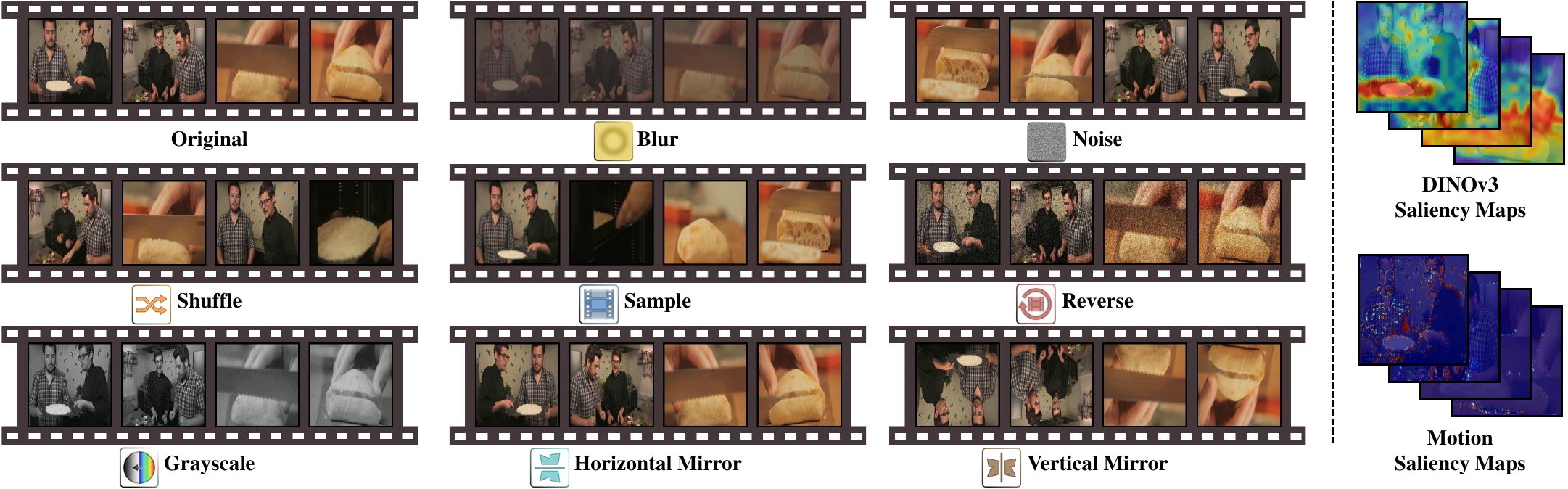}
\vspace{-0.5em}
\caption{Visual illustration of the eight video perturbation tools (left) and the two saliency maps (right). To demonstrate the diverse effects of our perturbation, we extract representative frames from the same timestamp across the original and perturbed video sequences. These tools encompass both temporal disruptions (Reverse, Shuffle, Sample) and spatial/frame-level distortions (Blur, Noise, Grayscale, Mirror). The right panel showcases the DINOv3-based spatial saliency maps and optical flow-based motion saliency maps.}
\label{fig_tool_sample}
\end{figure*}

\subsection{Details of Adaptive Perturbation Controller}\label{appendix_apc}
\alg \ref{alg:apc} defines the execution logic of the Adaptive Perturbation Controller, which transforms the raw video into a semantically-targeted negative sample. To facilitate precise tool selection, we introduce two specialized learnable embeddings:

\begin{itemize}
    \item Type Embeddings ($\mathbf{E}_{type}$): These embeddings are added to the projected hidden states of both the video-query context ($\mathbf{E}_{type}^{vq}$) and the tool descriptions ($\mathbf{E}_{type}^{tool}$). This allows the cross-attention mechanism to explicitly distinguish between the semantic “request” (context) and the “resource” (available tools), improving the alignment accuracy.

    \item ID Embeddings ($\mathbf{E}_{id}$): Each of the eight perturbation tools is assigned a unique learnable ID embedding. These are injected into the tool description features to ensure that even if two tools have semantically similar descriptions, the APC can maintain distinct routing preferences and ranking scores for each.
\end{itemize}

\begin{algorithm}[htbp]
\caption{Algorithm workflow of APC}
\label{alg:apc}
\begin{algorithmic}[1]
    \REQUIRE Raw Video $V$; Query $T$; Hidden States $\mathbf{H}$; Tool Description Embeddings $\{\mathbf{D}_i\}_{i=1}^8$; Threshold $\gamma$.
    \ENSURE Perturbed Video $V^{-}$; Negative Logit $q_t^n$.
    
    \STATE $\mathbf{H}' \leftarrow \text{Proj}(\mathbf{H}) + \mathbf{E}_{type}^{vq}$ \quad \COMMENT{\small Apply projection \& type embedding}
    \FOR{each tool $i \in \{1, \dots, 8\}$}
        \STATE $\mathbf{D}_i' \leftarrow \text{Proj}(\mathbf{D}_i) + \mathbf{E}_{type}^{tool} + \mathbf{E}_{id}^{i}$ \quad \COMMENT{\small Incorporate type \& ID embeddings}
    \ENDFOR

    \STATE $\mathbf{Q}_{cond} \leftarrow \text{Cross-Attn}(\mathbf{Q}, \mathbf{H}')$ \quad \COMMENT{\small Stage A: Contextual conditioning}
    \FOR{each tool $i \in \{1, \dots, 8\}$}
        \STATE $\mathbf{Q}_{route}^{i} \leftarrow \text{Cross-Attn}(\mathbf{Q}_{cond}, \mathbf{D}_i')$ \quad \COMMENT{\small Stage B: Tool-specific routing}
        \STATE $\mathbf{r}_i \leftarrow \text{Mean}(\mathbf{Q}_{route}^{i})$ \quad \COMMENT{\small Token aggregation}
        \STATE $s_i \leftarrow \text{ScoringHead}(\mathbf{r}_i)$ \quad \COMMENT{\small Generate logit}
    \ENDFOR

    \STATE $\mathbf{P} \leftarrow \text{Softmax}(\mathbf{s})$ \quad \COMMENT{\small Probability distribution over tools}
    \STATE $\mathcal{I}_{sorted} \leftarrow \text{Argsort}(\mathbf{P}, \text{descending})$ \quad \COMMENT{\small Rank by relevance}
    \STATE Find $k$ such that $\sum_{j=1}^{k} \mathbf{P}_{j} \geq \gamma$ \quad \COMMENT{\small Cumulative thresholding}
    \STATE $\mathcal{T}_{select} \leftarrow \{ \tau_{i} \mid i \in \mathcal{I}_{sorted}[1:k] \}$
    \STATE $V^{-} \leftarrow \mathcal{T}_{select}(V)$ \quad \COMMENT{\small Apply selected perturbation}
    \STATE $q_t^n \leftarrow \text{logit}_{\theta} (y_t \mid V^{-}, T, y_{<t})$ \quad \COMMENT{\small Obtain negative logit}
\end{algorithmic}
\end{algorithm}

\subsection{Details of Saliency-Guided Enhancement}\label{appendix_sge}

\paragraph{Spatial Saliency Extraction.} 
We utilize DINOv3 \cite{simeoni2025dinov3} as a spatial saliency extractor to capture foreground object importance. In DINOv3, the token sequence for each frame is structured as $\mathbf{X}_{DINO} = [\mathbf{x}_{cls}; \mathbf{x}_{reg}^{1 \dots 4}; \mathbf{x}_{p}^{5 \dots 5+N}]$, where $\mathbf{x}_{cls}$ is the [CLS] token, $\mathbf{x}_{reg}$ are 4 register tokens, and $\mathbf{x}_{p}$ are $N$ patch tokens. Following the self-attention mechanism of the Vision Transformer \cite{vaswani2017attention, dosovitskiy2021an}, we extract the attention weights from the last transformer layer. Specifically, we compute the attention from the [CLS] token to all patch tokens to identify region-wise significance. Let $\mathbf{A}^{(h)} \in \mathbb{R}^{(N+5) \times (N+5)}$ be the attention matrix for head $h_{D}$, the unified spatial saliency $\mathbf{s}_{spa}$ is obtained by averaging across all $H_{D}$ heads:
\begin{equation}
\mathbf{s}_{spa} = \text{softmax}\left(\frac{1}{H_{D}} \sum_{h_{D}=1}^{H_{D}} \mathbf{A}^{(h_{D})}_{[0, 5:N+5]}\right).
\end{equation}

\paragraph{Temporal Saliency Extraction.} 
To capture dynamic transitions, the motion saliency extractor estimates the optical flow field $\mathbf{F} \in \mathbb{R}^{H \times W \times 2}$ between consecutive frames.  We employ the Farneback algorithm \cite{farneback2003two} to derive the raw motion magnitude $\mathbf{M}$ at each pixel:\begin{equation}\mathbf{M} = \sqrt{F_x^2 + F_y^2},\end{equation}where $F_x$ and $F_y$ represent the horizontal and vertical displacement components of the flow field $\mathbf{F}$, respectively. While $\mathbf{M}$ captures all pixel-level changes, it is inherently susceptible to background noise and global camera jitter. To refine this into a meaningful saliency signal, we apply a dual-stage filtering process. First, a spatial Gaussian filter is used to smooth the magnitude map, reducing isolated pixel noise and emphasizing coherent object movement. Second, we implement a temporal band-pass filter across a sliding window of frames. This step is crucial as it suppresses low-frequency background drifts (\eg, slow camera panning) and high-frequency erratic noise, thereby isolating the mid-to-high frequency motion characteristic of intentional actions. The motion saliency map $\mathbf{s}_{mot}$ highlights spatiotemporal regions with significant dynamic content.

\paragraph{Dynamic Fusion.} As illustrated in \fig \ref{fig_framework} (c), the final spatiotemporal enhancement is modulated by a learnable gater. Since $\mathbf{s}_{mot}$ is derived at a pixel-level resolution while $\mathbf{s}_{spa}$ is extracted at the patch-level (DINOv3), these saliency maps exhibit heterogeneous spatial dimensions. To bridge this gap, we first perform individual normalization to ensure both signals are mapped into a unified $[0, 1]$ range. Subsequently, both $\mathbf{s}_{mot}$ and $\mathbf{s}_{spa}$ are aligned to the target visual patch size of the VLLM backbone (\eg, $16 \times 16$ for Qwen3-VL \cite{Qwen3-VL}) through bilinear interpolation. Conditioned on the video-query hidden states $\mathbf{H}$, the gating network employs a multi-layer perceptron (MLP) to learn a dynamic fusion weight $\beta \in [0, 1]$. The unified spatiotemporal weight is $\mathbf{w}_{sal} = \beta \cdot \mathbf{s}_{spa} + (1-\beta) \cdot \mathbf{s}_{mot}$. The positive vision tokens $\mathbf{X}_v'$ are then enhanced via element-wise reweighting: $\mathbf{X}_v' = \mathbf{w}_{sal} \odot \mathbf{X}_v$, where $\odot$ denotes the Hadamard product. The enhanced vison tokens $\mathbf{X}_v'$ are fed back into the LLM decoder along with the original text tokens $\mathbf{X}_t$. The saliency-guided positive logit $q_t^p$ is obtained:
\begin{equation}
q_t^p = \text{logit}_{\theta} (y_t \mid \mathbf{X}_v', \mathbf{X}_t, y_{<t}).
\end{equation}
This mechanism ensures that the positive pass is anchored to visual evidence that is both spatially salient and temporally active, effectively establishing a robust “upper bound” for grounded reasoning. \alg \ref{alg:sge} defines the execution logic of the Saliency-Guided Enhancement. To enhance the precision of the generated saliency maps, several specialized techniques are employed within the algorithm:
\begin{itemize}
    \item Adaptive Temporal Band-pass Filtering: To isolate meaningful motion from the raw magnitude $\mathbf{M}$, we implement a statistical filtering mechanism based on a sliding temporal window $W$.
    \begin{itemize}
        \item Low-frequency Suppression ($\lambda_{low}$): By subtracting $(\mu + \lambda_{low} \sigma)$, we eliminate low-magnitude motion components typically associated with background drift or slow camera panning.
        \item High-frequency Removal ($\lambda_{high}$): An indicator function $\mathbb{I}(\mathbf{M}_i \leq \mu + \lambda_{high} \sigma)$ is used to suppress erratic, high-frequency noise caused by video artifacts or sudden camera shakes.
    \end{itemize}
\end{itemize}

\begin{algorithm}[htbp]
\caption{Algorithm workflow of SGE}
\label{alg:sge}
\begin{algorithmic}[1]
    \REQUIRE Video Frames $\{V_i\}_{i=1}^K$; Query Tokens $\mathbf{X}_t$; Hidden States $\mathbf{H}$; Patch Size $P$.
    \ENSURE Enhanced Vision Tokens $\mathbf{X}_v'$; Positive Logit $q_p^t$.

    \STATE $\mathbf{A} \leftarrow \text{Extract-Attn}(\text{DINOv3}(V))$ \quad \COMMENT{\small Get CLS-to-patch attention}
    \STATE $\mathbf{s}_{spa} \leftarrow \text{Head-Agg}(\mathbf{A}_{[:, 0, 5:N+5]})$ \quad \COMMENT{\small Filter out register tokens}

    \FOR{each consecutive pair $(V_{i-1}, V_i)$}
        \STATE $\mathbf{F} \leftarrow \text{Farneback}(V_{i-1}^{gray}, V_i^{gray})$ \quad \COMMENT{\small Estimate optical flow field}
        \STATE $\mathbf{M}_i \leftarrow \sqrt{F_x^2 + F_y^2}$ \quad \COMMENT{\small Compute raw motion magnitude}
        \STATE $\mathbf{M}_i \leftarrow \text{Gaussian-Blur}(\mathbf{M}_i, \pi)$ \quad \COMMENT{\small Spatial smoothing}
    \ENDFOR
    
    \FOR{each frame $i$ in window $W$}
        \STATE $\mu, \sigma \leftarrow \text{Mean}(\mathbf{M}_{W}), \text{Std}(\mathbf{M}_{W})$ \quad \COMMENT{\small Compute temporal statistics}
        \STATE $\mathbf{s}_{mot}^i \leftarrow \max(\mathbf{M}_i - (\mu + \lambda_{low} \sigma), 0)$ \quad \COMMENT{\small Suppress background drift}
        \STATE $\mathbf{s}_{mot}^i \leftarrow \mathbf{s}_{mot}^i \cdot \mathbb{I}(\mathbf{M}_i \leq \mu + \lambda_{high} \sigma)$ \quad \COMMENT{\small Suppress high-freq noise}
    \ENDFOR

    \STATE $\bar{\mathbf{s}}_{spa}, \bar{\mathbf{s}}_{mot} \leftarrow \text{Normalize}(\mathbf{s}_{spa}, \mathbf{s}_{mot}) \in [0, 1]$
    \STATE $\hat{\mathbf{s}}_{spa}, \hat{\mathbf{s}}_{mot} \leftarrow \text{Interpolate}(\bar{\mathbf{s}}_{spa}, \bar{\mathbf{s}}_{mot}, P \times P)$ 
    \STATE $\beta \leftarrow \text{Sigmoid}(\text{MLP}(\text{Mean}(\mathbf{H})))$ \quad \COMMENT{\small Dynamic gating}
    \STATE $\mathbf{w}_{sal} \leftarrow \beta \cdot \hat{\mathbf{s}}_{spa} + (1-\beta) \cdot \hat{\mathbf{s}}_{mot}$ \quad \COMMENT{\small Weighted fusion}

    \STATE $\mathbf{X}_v' \leftarrow \mathbf{w}_{sal} \odot \mathbf{X}_v$ \quad \COMMENT{\small Element-wise token reweighting}
    \STATE $q_t^p \leftarrow \text{logit}_{\theta} (y_t \mid \mathbf{X}_v', \mathbf{X}_t, y_{<t})$ \quad \COMMENT{\small Generate positive logit}

\end{algorithmic}
\end{algorithm}

\subsection{Details of Optimization}\label{appendix_opt}
We formulate the combined coordination of perturbation selection and saliency fusion as a Reinforcement Learning (RL) \cite{sutton1998reinforcement, li2017deep} problem. The primary motivation for adopting an RL framework is that the APC tool selection is a non-differentiable operation. The discrete choice of specific perturbation tools from a dictionary $\mathcal{P}$ prevents the use of standard backpropagation \cite{rumelhart1986learning}. By interacting with the frozen VLLM, both the APC and SGE learn to maximize suppressing hallucination pathways.

\paragraph{APC Selection Policy (Discrete).} To accommodate the non-differentiable tool selection, we model the APC as a set of independent Bernoulli distributions \cite{casella2024statistical}. For each of the eight tools in $\mathcal{P}$, the router outputs a logit that is mapped via a sigmoid function to a selection probability. This Bernoulli formulation allows the model to sample multiple tools simultaneously, enabling the creation of complex negative samples that better capture compositional hallucinations.

\paragraph{SGE Fusion Policy (Continuous).} The fusion weight $\beta$ for spatiotemporal enhancement is modeled as a continuous action. During training, we sample $\beta$ from a Normal distribution $\mathcal{N}(\mu, \sigma^2)$, where the mean $\mu$ is provided by the Gater and $\sigma$ is a fixed standard deviation to facilitate exploration. During inference, the mean value $\mu$ is used deterministically.

\paragraph{Training Stability.} Given the high cost of VLLM inference, we implement gradient accumulation across $b$ steps to stabilize the policy updates. This effectively simulates a larger batch size for the RL agent, ensuring that the learned policies are robust across diverse video contexts. Additionally, we apply gradient clipping to prevent policy collapse during the early stages of spatiotemporal exploration. The complete training procedure is detailed in \alg \ref{alg:optimization}.

\begin{algorithm}[htbp]
\caption{\benchmark{} Training workflow}
\label{alg:optimization}
\begin{algorithmic}[1]
    \REQUIRE Training set $\mathcal{D}$; Frozen VLLM $\theta$; Initialized Policy $\phi$; Baseline $B$; Accumulation Steps $b$; Prediction $y_{pred}$; Ground Truth $y_{gt}$; Decay Factor $\varphi$; Learning Rate $l$; Gradient Clip Threshold $\eta$.    \ENSURE Optimized Parameters $\phi$.

    \STATE $B \leftarrow 0, \quad step \leftarrow 0, \quad \nabla \mathcal{L} \leftarrow 0$
    
    \FOR{each batch $(y_{pred}, y_{gt}) \in \mathcal{D}$}
        \STATE $R \leftarrow 1.0$ \textbf{if} $y_{pred} = y_{gt}$ \textbf{else} $-1.0$ 
        \STATE $B \leftarrow \varphi \cdot B + (1-\varphi) \cdot R$ 
        \STATE $Adv \leftarrow R - B$ \quad \COMMENT{\small Calculate advantage signal}
        \STATE $\mathcal{L} \leftarrow - \left(\sum \log P(\mathbf{a}_{apc}) + \log P(\mathbf{a}_{sge})\right) \cdot Adv$ 
        \STATE $\nabla \mathcal{L} \leftarrow \nabla \mathcal{L} + \text{Backprop}(\mathcal{L}) / M$ \quad \COMMENT{\small Accumulate}
        \STATE $step \leftarrow step + 1$
        
        \IF{$step \pmod M = 0$}
            \STATE $\phi \leftarrow \text{OptimizerStep}(\phi, \nabla \mathcal{L}, l, \text{clip}=\eta)$ \quad \COMMENT{\small Update}
            \STATE $\nabla \mathcal{L} \leftarrow 0$
        \ENDIF
    \ENDFOR
\end{algorithmic}
\end{algorithm}

\section{Experimental Details}\label{Experimental Details}
\subsection{Model Details}\label{Models}
We evaluated 39 VLLMs from 13 different model families, including ten open-source model families:
Qwen3-VL \cite{Qwen3-VL}, 
Qwen2.5-VL \cite{bai2025qwen2_5}, 
LLaVA-NeXT-Video \cite{zhang2024llavanext-video},
VideoLLaMA3 \cite{damonlpsg2025videollama3},
GLM series \cite{vteam2025glm45vglm41vthinkingversatilemultimodal},
Molmo2 \cite{clark2026molmo2openweightsdata},
Kimi-VL \cite{kimiteam2025kimivltechnicalreport},
InternVL3.5 \cite{wang2025internvl3_5},
VideoChat-Flash \cite{li2024videochatflash},
MiniCPM-V series \cite{yu2025minicpmv45cookingefficient, yao2024minicpm},
and three close-source model families: 
GPT series \cite{openai2025gpt4o, openai2025gpt5, openai2025gpt5_2},
Gemini series \cite{google2025gemini2_5flash, google2025gemini2_5pro, google2025gemini3pro}, Doubao series \cite{doubao2025seed1_6, doubao2025seed1_8}. These models represent a wide variety of architectural designs and training paradigms. Additionally, we included a human baseline. All models and their checkpoints are listed in \tab \ref{tab_all_models}. The open-source models are available via HuggingFace (\url{https://huggingface.co/}), while the proprietary models are accessible through their respective providers’ APIs. All evaluations for the proprietary models were conducted in January 2026.

\begin{table*}[!t]
  \centering
  \caption{Details on model names and corresponding checkpoints.}
  \label{tab_all_models}
  \setlength{\tabcolsep}{1.5pt}
  \resizebox{0.98\textwidth}{!}{ 
  \begin{tabular}{lcll} 
    \toprule
    \textbf{Model} & \textbf{Size} & \textbf{Checkpoint} & \textbf{API} \\
    \midrule
    \multicolumn{4}{c}{\textit{Open-source Models}} \\
    \midrule
    \multirow{5}{*}{Qwen3-VL-Instruct} 
    & 235B & \url{https://huggingface.co/Qwen/Qwen3-VL-235B-A22B-Instruct} & \multirow{13}{*}{\url{https://www.aliyun.com}} \\
    & 32B & \url{https://huggingface.co/Qwen/Qwen3-VL-32B-Instruct} & \\
    & 8B & \url{https://huggingface.co/Qwen/Qwen3-VL-8B-Instruct} & \\
    & 4B & \url{https://huggingface.co/Qwen/Qwen3-VL-4B-Instruct} & \\
    & 2B & \url{https://huggingface.co/Qwen/Qwen3-VL-2B-Instruct} & \\
    \cmidrule(lr){1-3}
    \multirow{5}{*}{Qwen3-VL-Thinking} 
    & 235B & \url{https://huggingface.co/Qwen/Qwen3-VL-235B-A22B-Thinking} & \\
    & 32B & \url{https://huggingface.co/Qwen/Qwen3-VL-32B-Thinking} & \\
    & 8B & \url{https://huggingface.co/Qwen/Qwen3-VL-8B-Thinking} & \\
    & 4B & \url{https://huggingface.co/Qwen/Qwen3-VL-4B-Thinking} & \\
    & 2B & \url{https://huggingface.co/Qwen/Qwen3-VL-2B-Thinking} & \\
    \cmidrule(lr){1-3}
    \multirow{3}{*}{Qwen2.5-VL-Instruct} 
    & 32B & \url{https://huggingface.co/Qwen/Qwen2.5-VL-32B-Instruct} & \\
    & 7B & \url{https://huggingface.co/Qwen/Qwen2.5-VL-7B-Instruct} & \\
    & 3B & \url{https://huggingface.co/Qwen/Qwen2.5-VL-3B-Instruct} & \\
    \midrule
    \multirow{2}{*}{LLaVA-NeXT-Video} 
    & 34B & \url{https://huggingface.co/llava-hf/LLaVA-NeXT-Video-34B-hf} & \multirow{2}{*}{-} \\
    & 7B & \url{https://huggingface.co/llava-hf/LLaVA-NeXT-Video-7B-hf} & \\
    \midrule
    \multirow{2}{*}{VideoLLaMA3} 
    & 7B & \url{https://huggingface.co/DAMO-NLP-SG/VideoLLaMA3-7B} & \multirow{2}{*}{-} \\
    & 2B & \url{https://huggingface.co/DAMO-NLP-SG/VideoLLaMA3-2B} & \\
    \midrule
    GLM-4.6v & 108B & \url{https://huggingface.co/zai-org/GLM-4.6V} & \multirow{3}{*}{\url{https://bigmodel.cn}} \\
    GLM-4.6v-flash & 108B & \url{https://huggingface.co/zai-org/GLM-4.6V-Flash} & \\
    GLM-4.5v & 108B & \url{https://huggingface.co/zai-org/GLM-4.5V} & \\

    \midrule
    \multirow{2}{*}{Molmo2} 
    & 8B & \url{https://huggingface.co/allenai/Molmo2-8B} & \multirow{2}{*}{-} \\
    & 4B & \url{https://huggingface.co/allenai/Molmo2-4B} & \\
    \midrule
    Kimi-VL-Thinking & 16B & \url{https://huggingface.co/moonshotai/Kimi-VL-A3B-Thinking-2506} & \multirow{2}{*}{\url{https://platform.moonshot.ai}} \\
    Kimi-VL-Instruct & 16B & \url{https://huggingface.co/moonshotai/Kimi-VL-A3B-Instruct} & \\
    \midrule
    \multirow{3}{*}{InternVL3.5} 
    & 30B & \url{https://huggingface.co/OpenGVLab/InternVL3_5-30B-A3B} & \multirow{3}{*}{-} \\
    & 8B & \url{https://huggingface.co/OpenGVLab/InternVL3_5-8B} & \\
    & 4B & \url{https://huggingface.co/OpenGVLab/InternVL3_5-4B} & \\
    \midrule
    \multirow{2}{*}{VideoChat-Flash} 
    & 7B & \url{https://huggingface.co/OpenGVLab/VideoChat-Flash-Qwen2-7B_res448} & \multirow{2}{*}{-} \\
    & 2B & \url{https://huggingface.co/OpenGVLab/VideoChat-Flash-Qwen2_5-2B_res448} &  \\
    \midrule
    MiniCPM-V-4.5 & 8B & \url{https://huggingface.co/openbmb/MiniCPM-V-4_5} & \multirow{2}{*}{-}\\
    MiniCPM-V-4 & 4B & \url{https://huggingface.co/openbmb/MiniCPM-V-4} & \\
    \midrule
    \multicolumn{4}{c}{\textit{Proprietary Models}} \\
    \midrule
    GPT-5.2 & - & - & \multirow{3}{*}{\url{https://openai.com}} \\
    GPT-5 & - & - & \\
    GPT-4o & - & - & \\
    \midrule
    Gemini-3-pro & - & -  & \multirow{3}{*}{\url{https://ai.google.dev}} \\
    Gemini-2.5-pro & - & - & \\
    Gemini-2.5-flash & - & - & \\
    \midrule
    Doubao-Seed-1.8 & - & - & \multirow{2}{*}{\url{https://console.volcengine.com}} \\
    Doubao-Seed-1.6 & - & - & \\
    \bottomrule
  \end{tabular}
  }
\end{table*}

\subsection{Metrics Details}\label{Metrics Details}
We adopt Accuracy as the primary evaluation metric for both YNQA and MCQA tasks. It is formulated as:
\begin{equation}
Acc = \frac{1}{n} \sum_{i=1}^{n} \mathbb{I}(y_{pred}^i = y_{gt}^i)
\end{equation}
where $n$ denotes the total number of test queries. $y_{pred}^i$ and $y_{gt}^i$ represent the predicted and ground-truth labels respectively. $\mathbb{I}(\cdot)$ is the indicator function. This unified metric allows for a consistent and direct comparison across different testing types. We also provide the following granular metrics in \benchmark{}.

\paragraph{YNQA Subset Accuracy.}
To identify potential linguistic biases, such as a “Yes-bias”, we calculate accuracy independently for affirmative and negative queries: $YesAcc = \frac{1}{n_{yes}} \sum_{i \in \mathcal{I}_{yes}} \mathbb{I}(y_{pred}^i = y_{gt}^i)$, and $NoAcc = \frac{1}{n_{no}} \sum_{i \in \mathcal{I}_{no}} \mathbb{I}(y_{pred}^i = y_{gt}^i)$, where $n_{yes}$ and $n_{no}$ represent the number of queries where the ground truth is “Yes” and “No”, respectively.

\paragraph{MCQA Discriminative Metrics.}
For multiple-choice tasks, we employ Macro-averaged Precision, Recall, and F1-score to assess the model’s discriminative power across all choice categories. By treating each option (\eg, A, B, and C) as an independent class, these metrics prevent performance from being masked by simple option biases:
\begin{itemize}
    \item \textbf{Macro-Precision} ($P_{macro} = \frac{1}{K} \sum_{i=1}^{K} \frac{TP_i}{TP_i + FP_i}$): Measures the average exactness of the model in identifying the correct spatiotemporal description across all $K$ candidate categories.
    \item \textbf{Macro-Recall} ($R_{macro} = \frac{1}{K} \sum_{i=1}^{K} \frac{TP_i}{TP_i + FN_i}$): Evaluates the model’s average ability to correctly retrieve the grounded answer from each specific option pool, ensuring robust detection of adversarial distractors.
    \item \textbf{Macro-F1-Score} ($F1_{macro} = \frac{1}{K} \sum_{i=1}^{K} \frac{2 \cdot P_i \cdot R_i}{P_i + R_i}$): Provides a balanced harmonic mean of precision and recall across all classes, ensuring the evaluation is not skewed by the frequency of adversarial injections.
\end{itemize}

\subsection{Implementation Details}\label{Implementation_Details}
\begin{figure*}[htbp]
\centering
\includegraphics[width=0.90\linewidth]{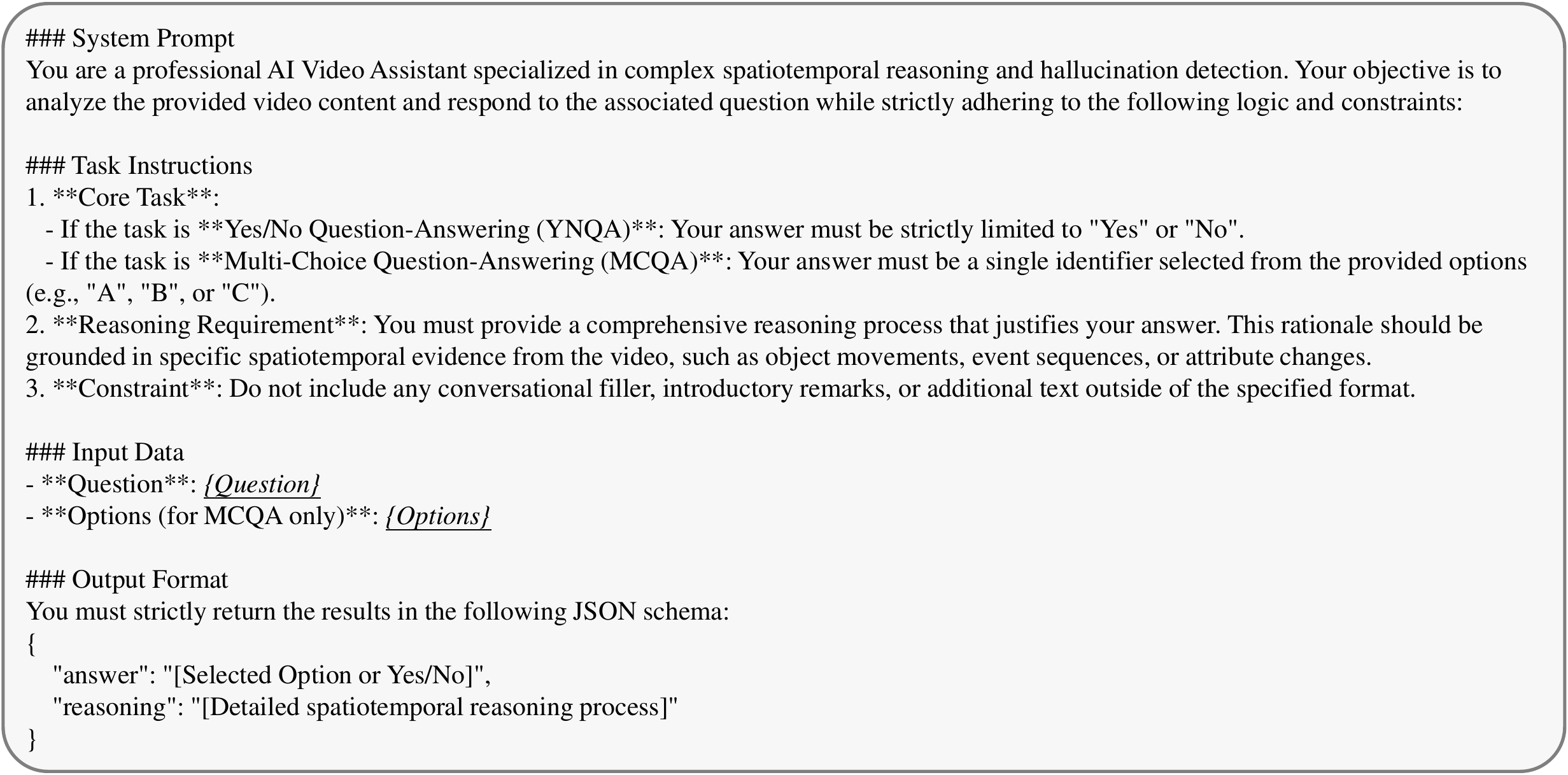}
\vspace{-0.5em}
\caption{Prompt for testing models . \underline{\textit{Terms}} (\eg, \underline{\textit{\{Question\}}} and \underline{\textit{\{Options\}}}) represent the dynamic input.}
\label{fig_qa_test_prompt}
\end{figure*}

All experiments were conducted using 8 NVIDIA RTX A6000 48GB GPUs. We adhered to the inference and model hyperparameters outlined in the respective original models, and employed greedy decoding during generation for a fair comparison. We utilized a unified prompt template throughout the experimental phase, as shown in \fig \ref{fig_qa_test_prompt}. To evaluate the efficacy of our proposed \approach{}, we repurpose the \benchmark{} benchmark into a supervised training and evaluation suite. Specifically, we partition the 823 videos and their corresponding 9,027 QA pairs into a 7:2:1 split for training, validation, and testing, respectively. This partition ensures that the performance gains are evaluated on unseen spatiotemporal scenarios. We implement \approach{} on two representative backbones: VideoLLaMA3-7B \cite{damonlpsg2025videollama3} and Qwen3-VL-Instruct-8B \cite{Qwen3-VL}, to demonstrate its scalability across different model capacities. We use a fixed predefined randomization seed across experiments. During the optimization phase of \approach{}, the VLLM backbone is kept strictly frozen, focusing solely on the lightweight APC and SGE modules. These are optimized using the REINFORCE algorithm with a learning rate of $l=1 \times 10^{-4}$. To stabilize the reinforcement learning process, we utilize a gradient accumulation of $b=32$ steps and a moving baseline with a decay factor of $\varphi =0.95$. For the APC module, the cumulative probability threshold for dynamic tool selection is set to $\gamma=0.4$. Spatial saliency in the SGE module is extracted using DINOv3 (dinov3-vitl16-pretrain-lvd1689m), while temporal motion features are computed via the Farneback algorithm with a standard window size $W$ of 5. The hyperparameter settings are shown in \tab \ref{tab_hyperparameters}.

\begin{table}[htbp]
  \centering
  \caption{Hyperparameters settings.}
  \label{tab_hyperparameters}
  \resizebox{0.48\textwidth}{!}{ 
  \begin{tabular}{lc} 
    \toprule
    \textbf{Hyperparameter} & \textbf{Value} \\
    \midrule
    Cumulative probability threshold ($\gamma$) & 0.4 \\
    Logit weight ($\alpha_1$) & 0.8 \\
    Logit weight ($\alpha_2$) & 0.4 \\
    Patch size ($P$) & 16/14 \\
    Standard deviation of spatial Gaussian filter ($\pi$) & 1.0 \\
    Temporal window size ($W$) & 5 \\
    Low-frequency threshold ($\lambda_{low}$) & 0.1 \\
    High-frequency threshold ($\lambda_{high}$) & 0.9 \\
    Standard deviation of distribution in SGE training ($\sigma$) & 0.1 \\
    Decay factor ($\varphi$) & 0.95 \\
    Gradient clip threshold ($\eta$) & 1.0 \\
    Learning rate ($l$) & $1 \times 10^{-4}$ \\
    Batch size ($b$) & 32 \\
    Random seed & 2025 \\
    \bottomrule
  \end{tabular}
  }
\end{table}

\subsection{Benchmark Results}\label{Results}
\begin{figure*}[!h]
\centering
\includegraphics[width=0.65\linewidth]{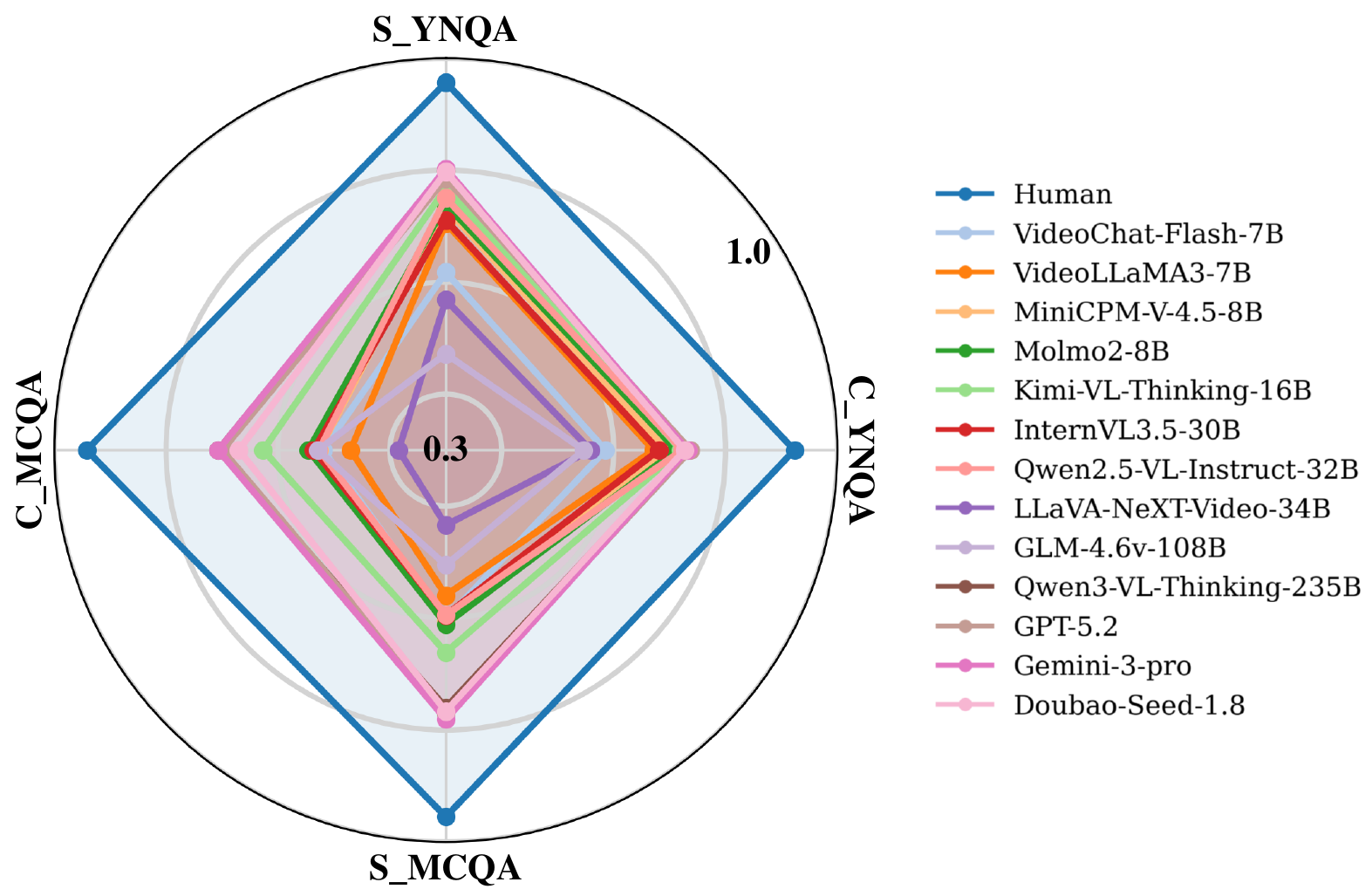}
\vspace{-0.5em}
\caption{Comparative accuracy radar chart of representative VLLMs across the four sub-tasks. The outer boundary represents human performance, highlighting the gap in spatiotemporal grounding for leading models from each architectural family.}
\label{small_radar}
\end{figure*}

\begin{figure*}[!h]
\centering
\includegraphics[width=0.85\linewidth]{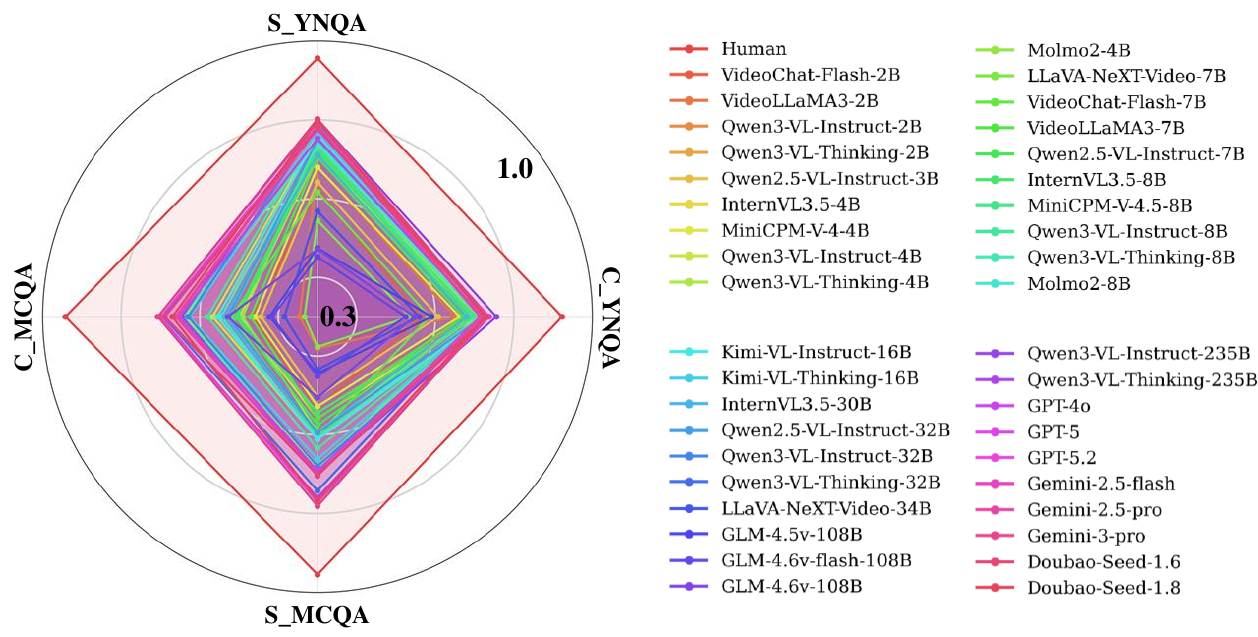}
\vspace{-0.5em}
\caption{Holistic accuracy distribution of all 39 evaluated models on \benchmark{}. The dense overlapping patterns reveal collective model vulnerabilities across simple and complex reasoning tasks relative to the near-perfect human baseline.}
\label{big_radar}
\end{figure*}

\tab \ref{tab_all_benchmark_results} provides the exhaustive accuracy for all evaluated VLLMs across the four sub-tasks in both real-world and AI-generated domains. To facilitate a deeper diagnostic of model behavior, \tab \ref{tab_all_real_results_all_metrics} and \tab \ref{tab_all_gen_results_all_metrics} report supplementary performance statistics, including class-specific accuracy for YNQA and Macro Precision, Recall, and F1-score for MCQA, under real-world and generated scenarios, respectively. These comprehensive tables offer a granular view of architectural robustness and potential linguistic biases across the heterogeneous video sources in \benchmark{}. Furthermore, \fig \ref{small_radar} and \fig \ref{big_radar} visualize the accuracy distribution across the four task dimensions, comparing representative model families and the full model suite, respectively, against the stable human performance ceiling.

\subsection{Baseline Methods}\label{Baseline Methods}
In this section, we provide a detailed description of baseline methods:

\begin{itemize}
    \item \textbf{MotionCD} \cite{kong2025mhbench}: This method utilizes motion-aware contrastive decoding to penalize hallucinations by corrupting temporal dynamics in the video. We follow the official implementation available at \url{https://github.com/Stevetich/EventHallusion}.
    \item \textbf{TCD} \cite{zhang2024eventhallusion}: This method focuses on mitigating event-level hallucinations by establishing a temporal contrastive baseline through frame shuffling. We follow the official implementation available at \url{https://github.com/xzhouzeng/MHBench}.
    \item \textbf{DINO-HEAL}  \cite{li2025vidhalluc}: This method employs spatial saliency extracted from DINOv2 features to enhance visual grounding and reduce hallucinations. We follow the official implementation available at \url{https://github.com/CyL97/VidHalluc}.
\end{itemize}

\subsection{Optimization and Efficiency}\label{Optimization and Efficiency}
To facilitate high-efficiency training, we pre-compute and locally store the tool embeddings and saliency maps, significantly reducing the redundant computational overhead during the RL phase. In the SGE module, saliency computation is strictly restricted to the frames selected by the VLLM's processor to minimize unnecessary visual processing. Furthermore, to maintain high throughput, \approach{} utilizes a parallel branching strategy where the SGE and APC modules independently process shared hidden states from the frozen backbone to generate positive and negative logit adjustments. These concurrent outputs are then integrated into a single contrastive decoding step, effectively mitigating hallucinations without introducing a sequential bottleneck or significantly increasing inference latency. We employ a dual-stage spatiotemporal perturbation pipeline: global temporal tools (\eg, temporal sampling) are first applied to the entire video sequence to disrupt long-range dynamics, followed by local spatial tools (\eg, grayscale transformation) that are restricted to the specific frames indexed by the VLLM's processor. This hierarchical approach ensures that negative samples incorporate both coarse-grained temporal inconsistencies and fine-grained spatial distortions while maintaining computational efficiency by avoiding redundant processing on unsampled frames.

As detailed in \tab \ref{tab_time_GB}, we evaluate the computational efficiency of \approach{} compared to existing baselines. Despite performing a dual-boundary calibration (APC and SGE), \approach{} maintains a manageable inference time. While sequential methods like MotionCD (8.05s) and TCD (7.95s) nearly double the backbone's latency (3.89s) due to independent dual-pass executions, our parallel branching strategy ensures that the positive and negative adjustments are computed concurrently from shared hidden states. This architectural choice effectively prevents a sequential bottleneck.

\begin{table}[htbp]
  \centering
  \caption{Inference efficiency on a single NVIDIA RTX A6000 (48GB) GPU. We report the average time per query for Qwen3-VL-Instruct-8B using different methods.}
  \label{tab_time_GB}
  \resizebox{0.5\textwidth}{!}{ 
  \begin{tabular}{lcc} 
    \toprule
    \textbf{Model} & \textbf{Average Time} & \textbf{Average Accuracy} \\
    \midrule
    Qwen3-VL-Instruct-8B & 3.89s & 0.67 \\
    +MotionCD & 8.05s & 0.68 \\
    +TCD & 7.95s & 0.69 \\
    +DINO-HEAL & 4.65s & 0.68 \\
    \approach{} & 9.12s & 0.73 \\
    \bottomrule
  \end{tabular}
  }
\end{table}

\begin{table}[htbp]
\centering
\caption{Hyperparameter sensitivity analysis of $\alpha_1$ and $\alpha_2$.}
\label{tab:hyper_sensitivity}
\resizebox{0.40\textwidth}{!}{ 
\begin{tabular}{c|cccccc}
\toprule
$\alpha_2 \setminus \alpha_1$ & 0.0 & 0.2 & 0.4 & 0.6 & 0.8 & 1.0 \\
\midrule
0.0 & 0.67 & 0.67 & 0.68 & 0.68 & 0.69 & 0.68 \\
0.2 & 0.67 & 0.68 & 0.69 & 0.71 & 0.72 & 0.69 \\
0.4 & 0.68 & 0.68 & 0.71 & 0.73 & \textbf{0.73} & 0.72 \\
0.6 & 0.69 & 0.69 & 0.70 & 0.72 & 0.72 & 0.71 \\
0.8 & 0.69 & 0.70 & 0.70 & 0.71 & 0.72 & 0.70 \\
1.0 & 0.71 & 0.70 & 0.69 & 0.70 & 0.70 & 0.69 \\
\bottomrule
\end{tabular}
}
\end{table}

\subsection{Hyperparameter Analysis}\label{Hyperparameter Analysis}
This additive refinement structure provides a robust theoretical foundation for multi-source logit calibration by decoupling the dual forces of evidence enhancement and bias suppression. Specifically, the positive grounding residual $(q_t^p - q_t^o)$ isolates the semantic gain achieved by focusing the visual encoder on spatiotemporal regions identified as salient. It effectively rewards tokens supported by verified visual evidence to correct for oversight errors in the original pass. Simultaneously, the negative suppression residual $(q_t^o - q_t^n)$ captures the pure hallucination bias that is the disparity between the model's standard reasoning and its over-reliance on linguistic priors. By pivoting these residuals around the baseline $q_t^o$, the formulation enables the independent calibration of grounding intensity ($\alpha_1$) and suppression strength ($\alpha_2$). This flexibility ensures that \approach{} can be fine-tuned to the specific inductive biases of different VLLM backbones, maintaining a stable decision boundary even in complex video scenarios.

As shown in \tab \ref{fig_hyper} confirms that the optimal balance is achieved at $\alpha_1 = 0.8$ and $\alpha_2 = 0.4$.

\subsection{Case Study}\label{Case Study}
To qualitatively evaluate the effectiveness of \approach{}, we present two representative cases.

\begin{figure*}[htbp]
\centering
\includegraphics[width=0.85\linewidth]{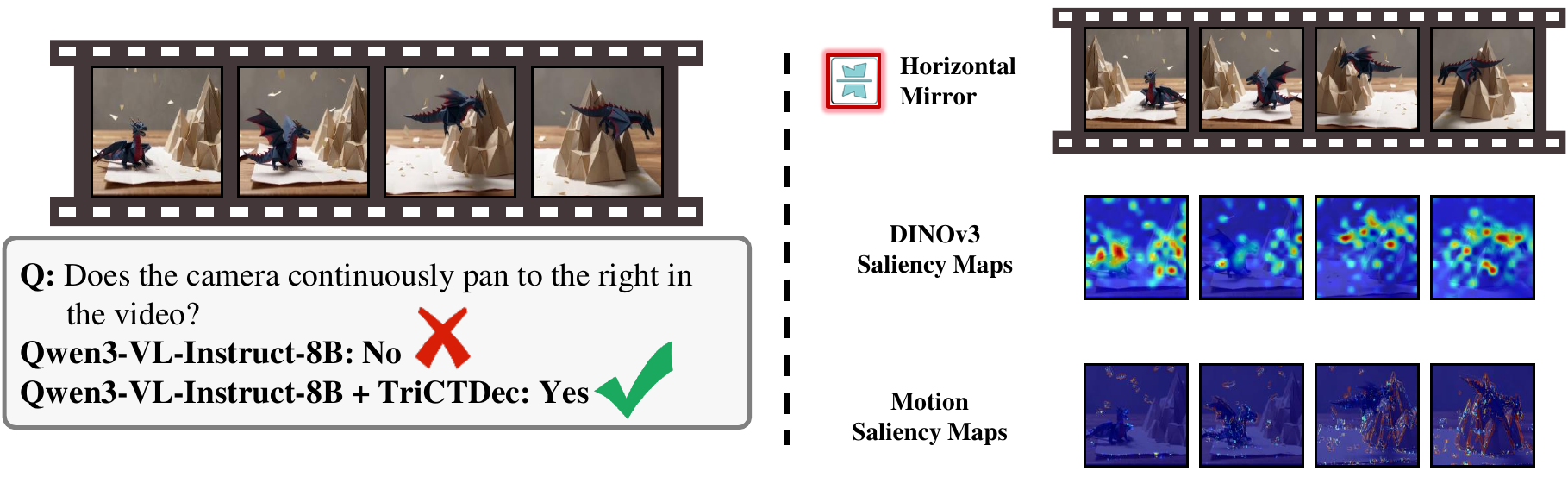}
\vspace{-0.5em}
\caption{Visualization of \approach{}'s robustness in the YNQA task.}
\label{case1}
\end{figure*}

\begin{figure*}[htbp]
\centering
\includegraphics[width=0.90\linewidth]{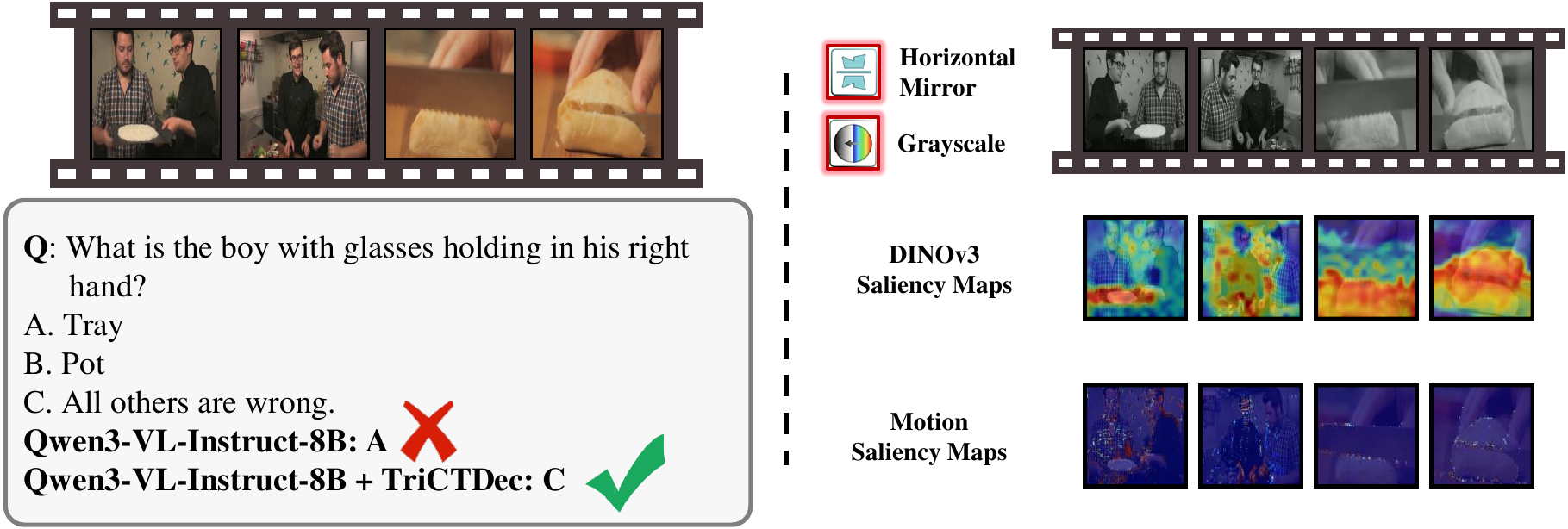}
\vspace{-0.5em}
\caption{Visualization of \approach{}'s robustness in the MCQA task.}
\label{case2}
\end{figure*}

In the YNQA task (\fig \ref{case1}), the base model fails to perceive the global camera motion, providing an incorrect “No” response. By contrast, \approach{} correctly identifies the panning direction. Similarly, in the MCQA task (\fig \ref{case2}), the backbone incorrectly hallucinates a “Tray” in the boy's hand. \approach{} successfully avoids this distractor and selects the adversarial option “C. All others are wrong”. This demonstrates that the synergy between our adaptive perturbations and dual-saliency grounding, incorporating both DINOv3 spatial features and motion cues, enables the model to maintain strict visual grounding even in complex, cluttered real-world environments.
\vspace{-1.5em}

\clearpage

\begin{table*}[p]
  \centering
  \caption{Main benchmark results on accuracy. The gray shaded cells highlight the \colorbox{best}{best} and \colorbox{second}{second-best} results in the open-source and commercial models, respectively. The \textit{\textbf{best model result}} among all models is highlighted. Avg means average.}
  \label{tab_all_benchmark_results}
  \setlength{\tabcolsep}{1.5pt}
  \resizebox{0.98\textwidth}{!}{ 
  \begin{tabular}{lcccccccccccc} 
    \toprule
    \multirow{2}{*}{\textbf{Model}} & \multirow{2}{*}{\textbf{Size}} & \multicolumn{5}{c}{\textbf{Real-world}} & \multicolumn{5}{c}{\textbf{AI-generated}} & \multirow{2}{*}{\textbf{Avg}} \\
    \cmidrule(lr){3-7} \cmidrule(lr){8-12}
    & & S\_YNQA & C\_YNQA & S\_MCQA & C\_MCQA & Avg & S\_YNQA & C\_YNQA & S\_MCQA & C\_MCQA & Avg & \\ 
    \midrule
Human&-& \textbf{0.96} & \textbf{0.93} & \textbf{0.95} & \textbf{0.95} & \textbf{0.95} & \textbf{0.96} & \textbf{0.91} & \textbf{0.96} & \textbf{0.93} & \textbf{0.94} & \textbf{0.95} \\
\midrule
\multicolumn{11}{c}{\textit{Open-source Models}} \\
\midrule
VideoChat-Flash \cite{li2024videochatflash} & 2B & 0.61 & 0.56 & 0.53 & 0.45 & 0.55 & 0.64 & 0.57 & 0.54 & 0.48 & 0.57 & 0.56 \\
VideoLLaMA3 \cite{damonlpsg2025videollama3} & 2B & 0.62 & 0.63 & 0.36 & 0.32 & 0.51 & 0.73 & 0.67 & 0.40 & 0.37 & 0.58 & 0.54 \\
Qwen3-VL-Instruct \cite{Qwen3-VL} & 2B & 0.68 & 0.66 & 0.50 & 0.46 & 0.59 & 0.75 & 0.71 & 0.59 & 0.54 & 0.67 & 0.63 \\
Qwen3-VL-Thinking \cite{Qwen3-VL} & 2B & 0.66 & 0.64 & 0.59 & 0.55 & 0.62 & 0.75 & 0.72 & 0.64 & 0.56 & 0.69 & 0.65 \\
Qwen2.5-VL-Instruct \cite{bai2025qwen2_5} & 3B & 0.59 & 0.60 & 0.45 & 0.40 & 0.52 & 0.68 & 0.62 & 0.56 & 0.49 & 0.61 & 0.57 \\
InternVL3.5 \cite{wang2025internvl3_5} & 4B & 0.67 & 0.67 & 0.55 & 0.46 & 0.61 & 0.74 & 0.72 & 0.56 & 0.53 & 0.66 & 0.64 \\
MiniCPM-V-4 \cite{yu2025minicpmv45cookingefficient} & 4B & 0.64 & 0.64 & 0.52 & 0.44 & 0.57 & 0.71 & 0.68 & 0.54 & 0.48 & 0.63 & 0.60 \\
Qwen3-VL-Instruct \cite{Qwen3-VL} & 4B & 0.71 & 0.66 & 0.59 & 0.55 & 0.64 & 0.81 & 0.77 & 0.64 & 0.58 & 0.72 & 0.68 \\
Qwen3-VL-Thinking \cite{Qwen3-VL} & 4B & 0.69 & 0.67 & 0.67 & 0.63 & 0.67 & 0.79 & 0.74 & 0.69 & 0.65 & 0.73 & 0.70 \\
Molmo2 \cite{clark2026molmo2openweightsdata} & 4B & 0.70 & 0.68 & 0.59 & 0.47 & 0.63 & 0.78 & 0.72 & 0.59 & 0.51 & 0.68 & 0.66 \\
LLaVA-NeXT-Video \cite{zhang2024llavanext-video} & 7B & 0.52 & 0.55 & 0.33 & 0.29 & 0.44 & 0.57 & 0.52 & 0.42 & 0.38 & 0.49 & 0.47 \\
VideoChat-Flash \cite{li2024videochatflash} & 7B & 0.61 & 0.59 & 0.58 & 0.49 & 0.58 & 0.62 & 0.58 & 0.62 & 0.55 & 0.60 & 0.59 \\
VideoLLaMA3 \cite{damonlpsg2025videollama3} & 7B & 0.65 & 0.64 & 0.55 & 0.46 & 0.59 & 0.75 & 0.70 & 0.57 & 0.48 & 0.65 & 0.62 \\
Qwen2.5-VL-Instruct \cite{bai2025qwen2_5} & 7B & 0.67 & 0.62 & 0.56 & 0.49 & 0.60 & 0.77 & 0.75 & 0.59 & 0.52 & 0.69 & 0.64 \\
InternVL3.5 \cite{wang2025internvl3_5} & 8B & 0.67 & 0.68 & 0.52 & 0.48 & 0.60 & 0.75 & 0.73 & 0.56 & 0.52 & 0.67 & 0.64 \\
MiniCPM-V-4.5 \cite{yu2025minicpmv45cookingefficient} & 8B & 0.69 & 0.69 & 0.59 & 0.51 & 0.64 & 0.76 & 0.70 & 0.62 & 0.54 & 0.68 & 0.66 \\
Qwen3-VL-Instruct \cite{Qwen3-VL} & 8B & 0.72 & 0.67 & 0.60 & 0.55 & 0.65 & 0.82 & 0.77 & 0.66 & 0.61 & 0.74 & 0.70 \\
Qwen3-VL-Thinking \cite{Qwen3-VL} & 8B & 0.68 & 0.68 & 0.65 & 0.64 & 0.67 & 0.79 & 0.76 & 0.71 & 0.65 & 0.74 & 0.71 \\
Molmo2 \cite{clark2026molmo2openweightsdata} & 8B & 0.70 & 0.69 & 0.60 & 0.53 & 0.65 & 0.77 & 0.73 & 0.62 & 0.56 & 0.69 & 0.67 \\
Kimi-VL-Instruct \cite{kimiteam2025kimivltechnicalreport} & 16B & 0.70 & 0.68 & 0.60 & 0.54 & 0.64 & 0.81 & \cellcolor{second} 0.78 & 0.63 & 0.57 & 0.72 & 0.68 \\
Kimi-VL-Thinking \cite{kimiteam2025kimivltechnicalreport} & 16B & 0.74 & 0.69 & 0.64 & 0.62 & 0.68 & 0.79 & 0.75 & 0.68 & 0.63 & 0.73 & 0.71 \\
InternVL3.5 \cite{wang2025internvl3_5} & 30B & 0.67 & 0.66 & 0.59 & 0.53 & 0.62 & 0.74 & 0.70 & 0.60 & 0.54 & 0.67 & 0.65 \\
Qwen2.5-VL-Instruct \cite{bai2025qwen2_5} & 32B & 0.69 & 0.69 & 0.55 & 0.50 & 0.62 & 0.79 & 0.75 & 0.64 & 0.55 & 0.71 & 0.67 \\
Qwen3-VL-Instruct \cite{Qwen3-VL} & 32B & 0.73 & \cellcolor{second} 0.70 & 0.66 & 0.62 & 0.69 & \cellcolor{second} 0.84 & 0.77 & 0.7 & 0.64 & 0.76 & 0.72 \\
Qwen3-VL-Thinking \cite{Qwen3-VL} & 32B & 0.73 & 0.67 & \cellcolor{second} 0.73 & \cellcolor{second} 0.70 & \cellcolor{second} 0.71 & 0.82 & 0.76 & \cellcolor{second} 0.76 & \cellcolor{second} 0.68 & \cellcolor{second} 0.77 & \cellcolor{second} 0.74 \\
LLaVA-NeXT-Video\cite{zhang2024llavanext-video} & 34B & 0.54 & 0.58 & 0.38 & 0.33 & 0.47 & 0.59 & 0.54 & 0.49 & 0.44 & 0.53 & 0.50 \\
GLM-4.5v \cite{vteam2025glm45vglm41vthinkingversatilemultimodal} & 108B & 0.46 & 0.59 & 0.43 & 0.39 & 0.47 & 0.48 & 0.59 & 0.46 & 0.45 & 0.50 & 0.49 \\
GLM-4.6v-flash \cite{vteam2025glm45vglm41vthinkingversatilemultimodal} & 108B & 0.43 & 0.53 & 0.42 & 0.35 & 0.44 & 0.47 & 0.52 & 0.49 & 0.48 & 0.49 & 0.46 \\
GLM-4.6v \cite{vteam2025glm45vglm41vthinkingversatilemultimodal} & 108B & 0.46 & 0.54 & 0.47 & 0.51 & 0.49 & 0.48 & 0.55 & 0.54 & 0.55 & 0.52 & 0.51 \\
Qwen3-VL-Instruct \cite{Qwen3-VL} & 235B & \cellcolor{second} 0.75 & \cellcolor{best} \textit{\textbf{0.73}} & 0.69 & 0.64 & 0.71 & \cellcolor{best} \textit{\textbf{0.84}} & \cellcolor{best} \textit{\textbf{0.79}} & 0.71 & 0.65 & 0.77 & 0.74 \\
Qwen3-VL-Thinking \cite{Qwen3-VL} & 235B & \cellcolor{best} 0.75 & 0.70 & \cellcolor{best} 0.75 & \cellcolor{best} 0.73 & \cellcolor{best} 0.73 & 0.84 & 0.77 & \cellcolor{best} 0.77 & \cellcolor{best} \textit{\textbf{0.68}} & \cellcolor{best} \textit{\textbf{0.78}} & \cellcolor{best} 0.76 \\
\midrule
\multicolumn{11}{c}{\textit{Proprietary Models}} \\
\midrule
GPT-4o \cite{openai2025gpt4o} & - & 0.72 & 0.69 & 0.71 & 0.66 & 0.70 & 0.78 & 0.76 & 0.67 & 0.64 & 0.73 & 0.71 \\
GPT-5 \cite{openai2025gpt5} & - & 0.72 & \cellcolor{best} 0.71 & 0.70 & \cellcolor{second} 0.72 & 0.72 & 0.81 & 0.76 & 0.70 & 0.66 & 0.75 & 0.73 \\
GPT-5.2 \cite{openai2025gpt5_2} & - & 0.74 & \cellcolor{second} 0.71 & \cellcolor{second} 0.77 & 0.72 & 0.74 & 0.82 & \cellcolor{best} 0.77 & 0.77 & \cellcolor{best} 0.68 & \cellcolor{best} 0.77 & \cellcolor{second} 0.76 \\
Gemini-2.5-flash \cite{google2025gemini2_5flash} & - & 0.74 & 0.68 & 0.70 & 0.69 & 0.71 & 0.82 & 0.76 & 0.70 & 0.64 & 0.75 & 0.73 \\
Gemini-2.5-pro \cite{google2025gemini2_5pro} & - & 0.74 & 0.68 & 0.71 & 0.68 & 0.71 & 0.81 & 0.75 & 0.70 & 0.64 & 0.74 & 0.73 \\
Gemini-3-pro \cite{google2025gemini3pro} & - & \cellcolor{best} \textit{\textbf{0.78}} & \cellcolor{second} 0.71 & \cellcolor{best} \textit{\textbf{0.79}} & \cellcolor{best} \textit{\textbf{0.74}} & \cellcolor{best} \textit{\textbf{0.76}} & \cellcolor{second} 0.82 & 0.76 & \cellcolor{best} \textit{\textbf{0.77}} & \cellcolor{second} 0.67 & \cellcolor{second} 0.77 & \cellcolor{best} \textit{\textbf{0.77}} \\
Doubao-Seed-1.6 \cite{doubao2025seed1_6} & - & 0.73 & 0.70 & 0.71 & 0.71 & 0.71 & 0.82 & \cellcolor{second} 0.76 & 0.69 & 0.67 & 0.75 & 0.73 \\
Doubao-Seed-1.8 \cite{doubao2025seed1_8} & - & \cellcolor{second} 0.76 & 0.70 & 0.76 & 0.67 & \cellcolor{second} 0.73 & \cellcolor{best} 0.82 & 0.76 & \cellcolor{second} 0.77 & 0.67 & 0.77 & 0.75 \\
    \bottomrule
  \end{tabular}
  }
\end{table*}

\clearpage

\begin{table*}[p]
\centering
\caption{Main results on other metrics under the real-word scenario.}
\label{tab_all_real_results_all_metrics}
\setlength{\tabcolsep}{1.5pt}
\resizebox{0.85\textwidth}{!}{ 
\begin{tabular}{lccccccccccc} 
\toprule
\multirow{2}{*}{\textbf{Model}} & \multirow{2}{*}{\textbf{Size}} & \multicolumn{2}{c}{\textbf{S\_YNQA}} & \multicolumn{2}{c}{\textbf{C\_YNQA}} & \multicolumn{3}{c}{\textbf{S\_MCQA}} & \multicolumn{3}{c}{\textbf{C\_MCQA}} \\ 
\cmidrule(lr){3-4} \cmidrule(lr){5-6} \cmidrule(lr){7-9} \cmidrule(lr){10-12}
& & YesAcc & NoAcc & YesAcc & NoAcc & 
\begin{tabular}[c]{@{}c@{}}Macro\\ Precision\end{tabular} & 
\begin{tabular}[c]{@{}c@{}}Macro\\ Recall\end{tabular} & 
\begin{tabular}[c]{@{}c@{}}Macro\\ F1\end{tabular} & 
\begin{tabular}[c]{@{}c@{}}Macro\\ Precision\end{tabular} & 
\begin{tabular}[c]{@{}c@{}}Macro\\ Recall\end{tabular} & 
\begin{tabular}[c]{@{}c@{}}Macro\\ F1\end{tabular} \\
\midrule
\multicolumn{12}{c}{\textit{Open-source Models}} \\
\midrule
VideoChat-Flash \cite{li2024videochatflash} & 2B & 0.89 & 0.42 & 0.89 & 0.32 & 0.54 & 0.53 & 0.53 & 0.45 & 0.45 & 0.45 \\
VideoLLaMA3 \cite{damonlpsg2025videollama3} & 2B & 0.62 & 0.62 & 0.62 & 0.65 & 0.46 & 0.34 & 0.23 & 0.34 & 0.31 & 0.19 \\
Qwen3-VL-Instruct \cite{Qwen3-VL} & 2B & 0.78 & 0.62 & 0.68 & 0.63 & 0.51 & 0.49 & 0.49 & 0.47 & 0.45 & 0.45 \\
Qwen3-VL-Thinking \cite{Qwen3-VL} & 2B & 0.74 & 0.61 & 0.65 & 0.64 & 0.60 & 0.58 & 0.57 & 0.57 & 0.55 & 0.54 \\
Qwen2.5-VL-Instruct \cite{bai2025qwen2_5} & 3B & 0.88 & 0.40 & 0.87 & 0.40 & 0.54 & 0.43 & 0.41 & 0.37 & 0.40 & 0.46 \\
InternVL3.5 \cite{wang2025internvl3_5} & 4B & 0.70 & 0.65 & 0.59 & 0.73 & 0.56 & 0.55 & 0.55 & 0.46 & 0.46 & 0.46 \\
MiniCPM-V-4 \cite{yu2025minicpmv45cookingefficient} & 4B & 0.88 & 0.48 & 0.83 & 0.51 & 0.53 & 0.52 & 0.52 & 0.46 & 0.44 & 0.44 \\
Qwen3-VL-Instruct \cite{Qwen3-VL} & 4B & 0.63 & 0.77 & 0.48 & 0.73 & 0.59 & 0.58 & 0.58 & 0.56 & 0.55 & 0.55 \\
Qwen3-VL-Thinking \cite{Qwen3-VL} & 4B & 0.66 & 0.71 & 0.50 & 0.78 & 0.68 & 0.67 & 0.67 & 0.65 & 0.63 & 0.62 \\
Molmo2 \cite{clark2026molmo2openweightsdata} & 4B & 0.83 & 0.61 & 0.77 & 0.62 & 0.60 & 0.59 & 0.60 & 0.48 & 0.47 & 0.47 \\
LLaVA-NeXT-Video \cite{zhang2024llavanext-video} & 7B & 0.62 & 0.45 & 0.65 & 0.48 & 0.33 & 0.33 & 0.33 & 0.30 & 0.28 & 0.29 \\
VideoChat-Flash \cite{li2024videochatflash} & 7B & 0.94 & 0.40 & 0.90 & 0.36 & 0.58 & 0.58 & 0.58 & 0.49 & 0.49 & 0.49 \\
VideoLLaMA3 \cite{damonlpsg2025videollama3} & 7B & 0.77 & 0.57 & 0.71 & 0.59 & 0.55 & 0.55 & 0.55 & 0.47 & 0.46 & 0.46 \\
Qwen2.5-VL-Instruct \cite{bai2025qwen2_5} & 7B & 0.64 & 0.68 & 0.56 & 0.71 & 0.57 & 0.55 & 0.55 & 0.51 & 0.49 & 0.49 \\
InternVL3.5 \cite{wang2025internvl3_5} & 8B & 0.71 & 0.65 & 0.67 & 0.68 & 0.52 & 0.52 & 0.52 & 0.49 & 0.48 & 0.48 \\
MiniCPM-V-4.5 \cite{yu2025minicpmv45cookingefficient} & 8B & 0.77 & 0.64 & 0.80 & 0.62 & 0.60 & 0.59 & 0.59 & 0.51 & 0.51 & 0.51 \\
Qwen3-VL-Instruct \cite{Qwen3-VL} & 8B & 0.63 & 0.77 & 0.49 & 0.76 & 0.62 & 0.60 & 0.60 & 0.57 & 0.55 & 0.55 \\
Qwen3-VL-Thinking \cite{Qwen3-VL} & 8B & 0.67 & 0.68 & 0.54 & 0.78 & 0.66 & 0.64 & 0.64 & 0.65 & 0.64 & 0.63 \\
Molmo2 \cite{clark2026molmo2openweightsdata} & 8B & 0.81 & 0.63 & 0.77 & 0.64 & 0.61 & 0.60 & 0.60 & 0.54 & 0.53 & 0.53 \\
Kimi-VL-Instruct \cite{kimiteam2025kimivltechnicalreport} & 16B & 0.69 & 0.70 & 0.50 & 0.76 & 0.61 & 0.59 & 0.59 & 0.55 & 0.54 & 0.53 \\
Kimi-VL-Thinking \cite{kimiteam2025kimivltechnicalreport} & 16B & 0.65 & 0.79 & 0.55 & 0.79 & 0.65 & 0.64 & 0.63 & 0.64 & 0.62 & 0.62 \\
InternVL3.5 \cite{wang2025internvl3_5} & 30B & 0.86 & 0.54 & 0.78 & 0.58 & 0.60 & 0.59 & 0.59 & 0.54 & 0.53 & 0.53 \\
Qwen2.5-VL-Instruct \cite{bai2025qwen2_5} & 32B & 0.65 & 0.72 & 0.56 & 0.78 & 0.55 & 0.54 & 0.54 & 0.52 & 0.50 & 0.49 \\
Qwen3-VL-Instruct \cite{Qwen3-VL} & 32B & 0.72 & 0.73 & 0.58 & 0.78 & 0.67 & 0.66 & 0.66 & 0.64 & 0.62 & 0.62 \\
Qwen3-VL-Thinking \cite{Qwen3-VL} & 32B & 0.69 & 0.76 & 0.48 & 0.81 & 0.75 & 0.72 & 0.72 & 0.72 & 0.70 & 0.70 \\
LLaVA-NeXT-Video \cite{zhang2024llavanext-video} & 34B & 0.65 & 0.47 & 0.68 & 0.50 & 0.37 & 0.38 & 0.37 & 0.34 & 0.32 & 0.33 \\
GLM-4.5v \cite{vteam2025glm45vglm41vthinkingversatilemultimodal} & 108B & 0.76 & 0.26 & 0.74 & 0.48 & 0.47 & 0.41 & 0.37 & 0.44 & 0.39 & 0.36 \\
GLM-4.6v-flash \cite{vteam2025glm45vglm41vthinkingversatilemultimodal} & 108B & 0.87 & 0.13 & 0.81 & 0.32 & 0.43 & 0.41 & 0.39 & 0.38 & 0.35 & 0.34 \\
GLM-4.6v \cite{vteam2025glm45vglm41vthinkingversatilemultimodal} & 108B & 0.81 & 0.22 & 0.79 & 0.36 & 0.50 & 0.46 & 0.44 & 0.55 & 0.50 & 0.48 \\
Qwen3-VL-Instruct \cite{Qwen3-VL} & 235B & 0.74 & 0.76 & 0.60 & 0.81 & 0.70 & 0.68 & 0.68 & 0.66 & 0.64 & 0.64 \\
Qwen3-VL-Thinking \cite{Qwen3-VL} & 235B & 0.71 & 0.78 & 0.55 & 0.80 & 0.77 & 0.74 & 0.74 & 0.75 & 0.73 & 0.72 \\
\midrule
\multicolumn{12}{c}{\textit{Proprietary Models}} \\
\midrule
GPT-4o \cite{openai2025gpt4o} & - & 0.62 & 0.80 & 0.58 & 0.76 & 0.72 & 0.71 & 0.71 & 0.67 & 0.66 & 0.66 \\
GPT-5 \cite{openai2025gpt5} & - & 0.80 & 0.67 & 0.58 & 0.80 & 0.71 & 0.70 & 0.70 & 0.74 & 0.72 & 0.72 \\
GPT-5.2 \cite{openai2025gpt5_2} & - & 0.69 & 0.77 & 0.57 & 0.81 & 0.79 & 0.77 & 0.77 & 0.75 & 0.72 & 0.72 \\
Gemini-2.5-flash \cite{google2025gemini2_5flash} & - & 0.72 & 0.75 & 0.53 & 0.79 & 0.73 & 0.70 & 0.70 & 0.71 & 0.69 & 0.68 \\
Gemini-2.5-pro \cite{google2025gemini2_5pro} & - & 0.73 & 0.74 & 0.54 & 0.78 & 0.74 & 0.70 & 0.70 & 0.71 & 0.68 & 0.68 \\
Gemini-3-pro \cite{google2025gemini3pro} & - & 0.68 & 0.84 & 0.53 & 0.84 & 0.80 & 0.79 & 0.79 & 0.75 & 0.74 & 0.74 \\
Doubao-Seed-1.6 \cite{doubao2025seed1_6} & - & 0.80 & 0.69 & 0.56 & 0.79 & 0.72 & 0.71 & 0.71 & 0.73 & 0.71 & 0.71 \\
Doubao-Seed-1.8 \cite{doubao2025seed1_8} & - & 0.77 & 0.76 & 0.55 & 0.80 & 0.79 & 0.76 & 0.76 & 0.73 & 0.67 & 0.67 \\
\bottomrule
\end{tabular}
}
\end{table*}

\clearpage

\begin{table*}[p]
\centering
\caption{Main results on other metrics under the AI-generated scenario.}
\label{tab_all_gen_results_all_metrics}
\setlength{\tabcolsep}{1.5pt}
\resizebox{0.85\textwidth}{!}{ 
\begin{tabular}{lccccccccccc} 
\toprule
\multirow{2}{*}{\textbf{Model}} & \multirow{2}{*}{\textbf{Size}} & \multicolumn{2}{c}{\textbf{S\_YNQA}} & \multicolumn{2}{c}{\textbf{C\_YNQA}} & \multicolumn{3}{c}{\textbf{S\_MCQA}} & \multicolumn{3}{c}{\textbf{C\_MCQA}} \\ 
\cmidrule(lr){3-4} \cmidrule(lr){5-6} \cmidrule(lr){7-9} \cmidrule(lr){10-12}
& & YesAcc & NoAcc & YesAcc & NoAcc & 
\begin{tabular}[c]{@{}c@{}}Macro\\ Precision\end{tabular} & 
\begin{tabular}[c]{@{}c@{}}Macro\\ Recall\end{tabular} & 
\begin{tabular}[c]{@{}c@{}}Macro\\ F1\end{tabular} & 
\begin{tabular}[c]{@{}c@{}}Macro\\ Precision\end{tabular} & 
\begin{tabular}[c]{@{}c@{}}Macro\\ Recall\end{tabular} & 
\begin{tabular}[c]{@{}c@{}}Macro\\ F1\end{tabular} \\
\midrule
\multicolumn{12}{c}{\textit{Open-source Models}} \\
\midrule
VideoChat-Flash \cite{li2024videochatflash} & 2B & 0.91 & 0.47 & 0.91 & 0.38 & 0.55 & 0.54 & 0.54 & 0.50 & 0.48 & 0.47 \\
VideoLLaMA3 \cite{damonlpsg2025videollama3} & 2B & 0.69 & 0.75 & 0.69 & 0.65 & 0.51 & 0.35 & 0.25 & 0.42 & 0.32 & 0.20 \\
Qwen3-VL-Instruct \cite{Qwen3-VL} & 2B & 0.81 & 0.72 & 0.73 & 0.70 & 0.61 & 0.58 & 0.58 & 0.56 & 0.53 & 0.52 \\
Qwen3-VL-Thinking \cite{Qwen3-VL} & 2B & 0.74 & 0.75 & 0.75 & 0.71 & 0.65 & 0.63 & 0.63 & 0.54 & 0.56 & 0.59 \\
Qwen2.5-VL-Instruct \cite{bai2025qwen2_5} & 3B & 0.90 & 0.55 & 0.78 & 0.55 & 0.63 & 0.55 & 0.54 & 0.53 & 0.48 & 0.46 \\
InternVL3.5 \cite{wang2025internvl3_5} & 4B & 0.67 & 0.79 & 0.68 & 0.75 & 0.57 & 0.56 & 0.56 & 0.54 & 0.53 & 0.53 \\
MiniCPM-V-4 \cite{yu2025minicpmv45cookingefficient} & 4B & 0.87 & 0.61 & 0.83 & 0.60 & 0.54 & 0.54 & 0.54 & 0.49 & 0.47 & 0.46 \\
Qwen3-VL-Instruct \cite{Qwen3-VL} & 4B & 0.70 & 0.88 & 0.62 & 0.85 & 0.64 & 0.64 & 0.64 & 0.59 & 0.59 & 0.58 \\
Qwen3-VL-Thinking \cite{Qwen3-VL} & 4B & 0.72 & 0.84 & 0.63 & 0.80 & 0.70 & 0.69 & 0.68 & 0.67 & 0.65 & 0.63 \\
Molmo2 \cite{clark2026molmo2openweightsdata} & 4B & 0.80 & 0.76 & 0.73 & 0.71 & 0.60 & 0.59 & 0.59 & 0.52 & 0.51 & 0.50 \\
LLaVA-NeXT-Video \cite{zhang2024llavanext-video} & 7B & 0.67 & 0.50 & 0.61 & 0.47 & 0.42 & 0.42 & 0.42 & 0.38 & 0.36 & 0.35 \\
VideoChat-Flash \cite{li2024videochatflash} & 7B & 0.95 & 0.41 & 0.91 & 0.40 & 0.62 & 0.62 & 0.62 & 0.56 & 0.54 & 0.54 \\
VideoLLaMA3 \cite{damonlpsg2025videollama3} & 7B & 0.79 & 0.72 & 0.76 & 0.66 & 0.59 & 0.57 & 0.57 & 0.49 & 0.48 & 0.46 \\
Qwen2.5-VL-Instruct \cite{bai2025qwen2_5} & 7B & 0.67 & 0.83 & 0.71 & 0.78 & 0.61 & 0.59 & 0.59 & 0.54 & 0.52 & 0.51 \\
InternVL3.5 \cite{wang2025internvl3_5} & 8B & 0.71 & 0.77 & 0.76 & 0.71 & 0.57 & 0.57 & 0.56 & 0.53 & 0.53 & 0.51 \\
MiniCPM-V-4.5 \cite{yu2025minicpmv45cookingefficient} & 8B & 0.83 & 0.72 & 0.83 & 0.62 & 0.62 & 0.62 & 0.62 & 0.56 & 0.55 & 0.54 \\
Qwen3-VL-Instruct \cite{Qwen3-VL} & 8B & 0.69 & 0.91 & 0.61 & 0.86 & 0.68 & 0.66 & 0.66 & 0.63 & 0.61 & 0.60 \\
Qwen3-VL-Thinking \cite{Qwen3-VL} & 8B & 0.76 & 0.81 & 0.68 & 0.80 & 0.72 & 0.70 & 0.70 & 0.67 & 0.65 & 0.64 \\
Molmo2 \cite{clark2026molmo2openweightsdata} & 8B & 0.79 & 0.75 & 0.75 & 0.72 & 0.63 & 0.62 & 0.62 & 0.57 & 0.56 & 0.55 \\
Kimi-VL-Instruct \cite{kimiteam2025kimivltechnicalreport} & 16B & 0.70 & 0.88 & 0.63 & 0.86 & 0.63 & 0.63 & 0.63 & 0.58 & 0.57 & 0.56 \\
Kimi-VL-Thinking \cite{kimiteam2025kimivltechnicalreport} & 16B & 0.72 & 0.84 & 0.64 & 0.81 & 0.69 & 0.68 & 0.68 & 0.65 & 0.63 & 0.62 \\
InternVL3.5 \cite{wang2025internvl3_5} & 30B & 0.82 & 0.70 & 0.84 & 0.62 & 0.60 & 0.60 & 0.60 & 0.55 & 0.55 & 0.54 \\
Qwen2.5-VL-Instruct \cite{bai2025qwen2_5} & 32B & 0.73 & 0.83 & 0.70 & 0.78 & 0.66 & 0.64 & 0.64 & 0.59 & 0.54 & 0.53 \\
Qwen3-VL-Instruct \cite{Qwen3-VL} & 32B & 0.77 & 0.88 & 0.69 & 0.82 & 0.71 & 0.69 & 0.69 & 0.67 & 0.64 & 0.64 \\
Qwen3-VL-Thinking \cite{Qwen3-VL} & 32B & 0.69 & 0.90 & 0.58 & 0.86 & 0.77 & 0.76 & 0.75 & 0.71 & 0.68 & 0.68 \\
LLaVA-NeXT-Video \cite{zhang2024llavanext-video} & 34B & 0.70 & 0.52 & 0.64 & 0.49 & 0.49 & 0.48 & 0.48 & 0.44 & 0.42 & 0.41 \\
GLM-4.5v \cite{vteam2025glm45vglm41vthinkingversatilemultimodal} & 108B & 0.72 & 0.34 & 0.68 & 0.55 & 0.50 & 0.43 & 0.41 & 0.49 & 0.44 & 0.42 \\
GLM-4.6v-flash \cite{vteam2025glm45vglm41vthinkingversatilemultimodal} & 108B & 0.82 & 0.26 & 0.76 & 0.39 & 0.51 & 0.47 & 0.46 & 0.49 & 0.47 & 0.45 \\
GLM-4.6v \cite{vteam2025glm45vglm41vthinkingversatilemultimodal} & 108B & 0.75 & 0.32 & 0.74 & 0.45 & 0.56 & 0.52 & 0.52 & 0.57 & 0.53 & 0.53 \\
Qwen3-VL-Instruct \cite{Qwen3-VL} & 235B & 0.78 & 0.88 & 0.70 & 0.84 & 0.72 & 0.71 & 0.70 & 0.68 & 0.65 & 0.65 \\
Qwen3-VL-Thinking \cite{Qwen3-VL} & 235B & 0.77 & 0.90 & 0.69 & 0.81 & 0.78 & 0.77 & 0.77 & 0.71 & 0.68 & 0.68 \\
\midrule
\multicolumn{12}{c}{\textit{Proprietary Models}} \\
\midrule
GPT-4o \cite{openai2025gpt4o} & - & 0.67 & 0.84 & 0.69 & 0.80 & 0.68 & 0.67 & 0.67 & 0.66 & 0.65 & 0.64 \\
GPT-5 \cite{openai2025gpt5} & - & 0.79 & 0.83 & 0.60 & 0.85 & 0.71 & 0.69 & 0.69 & 0.67 & 0.66 & 0.65 \\
GPT-5.2 \cite{openai2025gpt5_2} & - & 0.73 & 0.88 & 0.65 & 0.83 & 0.79 & 0.76 & 0.76 & 0.72 & 0.68 & 0.67 \\
Gemini-2.5-flash \cite{google2025gemini2_5flash} & - & 0.70 & 0.89 & 0.60 & 0.85 & 0.72 & 0.69 & 0.69 & 0.67 & 0.64 & 0.62 \\
Gemini-2.5-pro \cite{google2025gemini2_5pro} & - & 0.69 & 0.89 & 0.58 & 0.84 & 0.72 & 0.69 & 0.69 & 0.68 & 0.64 & 0.62 \\
Gemini-3-pro \cite{google2025gemini3pro} & - & 0.69 & 0.90 & 0.61 & 0.84 & 0.78 & 0.77 & 0.77 & 0.70 & 0.67 & 0.67 \\
Doubao-Seed-1.6 \cite{doubao2025seed1_6} & - & 0.81 & 0.83 & 0.62 & 0.84 & 0.70 & 0.69 & 0.69 & 0.68 & 0.67 & 0.66 \\
Doubao-Seed-1.8 \cite{doubao2025seed1_8} & - & 0.74 & 0.87 & 0.60 & 0.84 & 0.80 & 0.76 & 0.77 & 0.71 & 0.67 & 0.66 \\
\bottomrule
\end{tabular}
}
\end{table*}

\clearpage

\clearpage

\end{document}